\documentclass[runningheads]{llncs}
\usepackage{graphicx}
\usepackage{amsmath,amssymb} % define this before the line numbering.
\usepackage{soul}
\usepackage{subfig}
\usepackage[dvipsnames]{xcolor}
\usepackage{tabularx} % in the preamble
\usepackage[pagebackref=true,breaklinks=true,colorlinks,bookmarks=false]{hyperref}
\usepackage{booktabs}
\usepackage{pifont}
\usepackage{bold-extra}
\usepackage{dirtytalk}
\usepackage{mathtools}
\usepackage[numbers,sort]{natbib} % sort citations
\newcommand*\samethanks[1][\value{footnote}]{\footnotemark[#1]}

\newcommand{\cmark}{\ding{51}}%
\newcommand{\xmark}{\ding{55}}%

\newcommand{\bsldict}{\textsc{BslDict}}
\newcommand{\bslonek}{\text{BSL-1K}}
\newcommand{\baselineDict}{$\text{I3D}^{\text{\bsldict}}$}
\newcommand{\baselineBbc}{$\text{I3D}^{\text{\bslonek}}$}
\newcommand{\baselineBbcDict}{$\text{I3D}^{\text{\bslonek},\text{\bsldict}}$}
\def\sepappendix{0} % whether separate appendix or not 0/1
\usepackage{multirow}
\usepackage{siunitx}

\begin{document}
\pagestyle{headings}
\mainmatter

\def\ACCV20SubNumber{176}  % Insert your submission number here

%===========================================================
\title{Watch, read and lookup: learning to spot \\
signs from multiple supervisors}
\titlerunning{Watch, read and lookup}

\author{Liliane Momeni\thanks{Equal contribution} \and G\"{u}l Varol\samethanks \and Samuel Albanie\samethanks \and \\ Triantafyllos Afouras \and Andrew Zisserman}
\authorrunning{Momeni, Varol, Albanie, Afouras, Zisserman}
\institute{
    Visual Geometry Group, University of Oxford, UK \\
    \email{
        \{liliane,gul,albanie,afourast,az\}@robots.ox.ac.uk
    }
}

\maketitle

%===========================================================

\begin{abstract} 
The focus of this work is \textit{sign spotting}---given a video of an isolated sign, our task is to identify \textit{whether} and \textit{where} it has been signed in a continuous, co-articulated sign language video. To achieve this sign spotting task, we train a model using multiple types of available supervision by: (1) \textit{watching} existing sparsely labelled footage; (2) \textit{reading} associated subtitles (readily available translations of the signed content) which provide additional \textit{weak-supervision}; (3) \textit{looking up} words (for which no co-articulated labelled examples are available) in visual sign language dictionaries to enable novel sign spotting. These three tasks are integrated into a unified learning framework using the principles of Noise Contrastive Estimation and Multiple Instance Learning.  We validate the effectiveness of our approach on low-shot sign spotting benchmarks. In addition, we contribute a machine-readable British Sign Language (BSL) dictionary dataset of isolated signs, \bsldict, to facilitate study of this task. The dataset, models and code are available at our project page\footnote{\url{https://www.robots.ox.ac.uk/~vgg/research/bsldict/}}.
\end{abstract}
\section{Introduction} \label{sec:intro}

The objective of this work is to develop a \textit{sign spotting} model that can identify and localise instances of signs
%that correspond to a particular keyword
within sequences of continuous sign language. Sign languages represent the natural means of communication for deaf communities~\cite{sutton-spence_woll_1999} and sign spotting has a broad range of practical applications. Examples include: indexing videos of signing content by keyword to enable content-based search; gathering diverse dictionaries of sign exemplars from unlabelled footage for linguistic study; automatic feedback for language students via an \say{auto-correct} tool (e.g. \say{did you mean this sign?}); making voice activated wake word devices accessible to deaf communities; and building sign language datasets by automatically labelling examples of signs.

The recent marriage of large-scale, labelled datasets with deep neural networks has produced considerable progress in audio~\cite{coucke2019efficient,veniat2019stochastic} and visual~\cite{Momeni20,stafylakis2018zero} keyword spotting in \textit{spoken languages}. However, a direct replication of these keyword spotting successes in sign language requires a commensurate quantity of labelled data (note that modern audiovisual spoken keyword spotting datasets contain millions of densely labelled examples~\cite{chung2017lip,afouras2018lrs3}). Large-scale corpora of continuous, co-articulated\footnote{\textit{Co-articulation} refers to changes in the appearance of the current sign due to neighbouring signs.} signing from TV broadcast data have recently been built~\cite{bsl1k2020}, but the labels accompanying this data are: (1) sparse,  and (2) cover a limited vocabulary.

\begin{figure}[t]
    \centering
    \includegraphics[width=\linewidth]{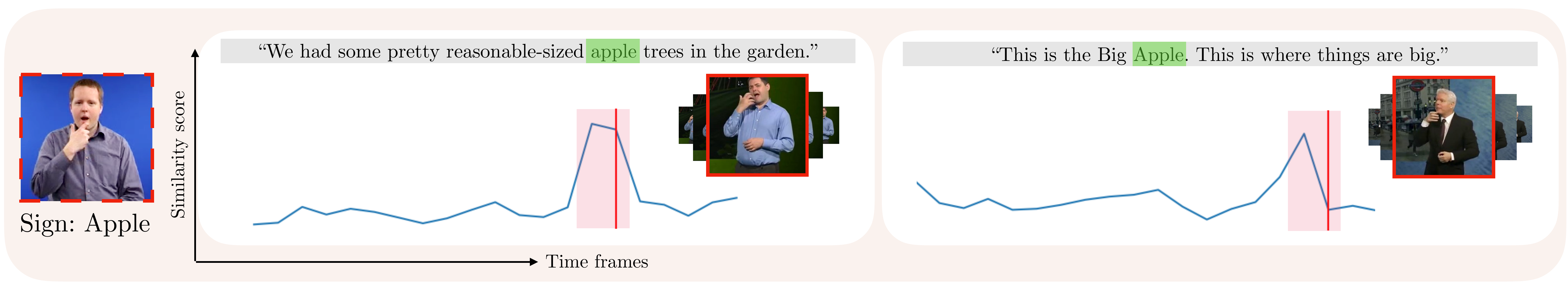}
    \caption{
    We consider the task of \textit{sign spotting} in co-articulated, continuous signing. Given a
    query dictionary video of
    an isolated sign (e.g. ``apple''),
    % keyword (e.g. ``apple'')
    we aim to identify \textit{whether} and \textit{where} it appears in videos of continuous signing.
    % by querying target footage with a learned representation of the dictionary entry for the keyword.
    The wide domain gap between dictionary examples of \textit{isolated} signs and target sequences of \textit{continuous} signing makes the task extremely challenging.
    }
    %\mbox{}\vspace{-0.8cm}\\
    \label{fig:kws-task}
\end{figure}

It might be thought that a sign language dictionary would offer a relatively straightforward solution to
the sign spotting task, particularly to the problem of covering only a limited vocabulary in existing large-scale corpora.
But, unfortunately, this is not the case due to the severe \textit{domain differences} between dictionaries 
and continuous signing in the wild.
The challenges are that sign language dictionaries typically: (i) 
consist of \textit{isolated signs} which differ in appearance from the
\textit{co-articulated} sequences of continuous signs (for which we ultimately
wish to perform spotting); and  (ii) differ in speed (are performed more slowly) relative to
co-articulated signing. 
Furthermore, (iii) dictionaries only  possess a few examples of each sign (so learning must be \textit{low shot});
and as one more challenge, (iv) there can be multiple signs corresponding to a single keyword, for example
due to regional variations of the sign language~\cite{bslcorpus17}.
We show through experiments in
Sec.~\ref{sec:experiments}, that directly training a sign spotter for
continuous signing on dictionary examples,
obtained from an internet-sourced sign
language dictionary, does indeed perform poorly.

To address these challenges, we 
propose
a unified framework in which sign spotting embeddings are learned
from the dictionary (to provide broad coverage of the lexicon) in combination with two additional sources of
supervision.  
In aggregate, these multiple types of supervision include: (1)
\textit{watching} sign language and learning from existing sparse
annotations; (2) exploiting
weak-supervision by \textit{reading} the subtitles that accompany the
footage and extracting candidates for signs that we expect to be
present; (3) \textit{looking up} words (for which we do not have
labelled examples) in a sign language dictionary.  The recent development of large-scale, subtitled corpora of continuous signing providing sparse annotations~\cite{bsl1k2020} allows us to study this problem setting directly.
We formulate our
approach as a Multiple Instance Learning problem in which positive
samples may arise from any of the three sources and employ Noise
Contrastive Estimation~\cite{gutmann2010noise} to learn a domain-invariant (valid across both isolated and co-articulated signing)
representation of signing content. 

We make the following six contributions: (1) We provide a machine readable British Sign Language (BSL) dictionary dataset of isolated signs, \bsldict, to facilitate study of the sign spotting task; (2) We propose a unified Multiple Instance Learning framework for learning sign embeddings suitable for spotting from three supervisory sources; (3) We validate the effectiveness of our approach on a co-articulated sign spotting benchmark for which only a small number (low-shot) of isolated signs are provided as labelled training examples, and (4) achieve state-of-the-art performance on the \bslonek{} sign spotting benchmark~\cite{bsl1k2020} (closed vocabulary). We show qualitatively that the learned embeddings can be used to (5) automatically mine new signing examples,  and (6) discover \say{faux amis} (false friends) between sign languages.

\section{Related Work}
\label{sec:related}

Our work relates to several themes in the literature: \textit{sign
language recognition} (and more specifically \textit{sign
spotting}), \textit{sign language datasets}, \textit{multiple
instance learning} and \textit{low-shot action localization}. We discuss each of these themes next. %\\

\noindent \textbf{Sign language recognition.}  
The study of automatic sign recognition has a rich history in the
computer vision community stretching back over 30 years, with early
methods developing carefully engineered features to model trajectories
and shape~\cite{Kadir04a,Tamura88,Starner95,Fillbrandt2003}. A series of
techniques then emerged which made effective use of hand and body pose
cues through robust keypoint estimation
encodings~\cite{Buehler09,Cooper2011,Ong2012,Pfister14}.  Sign
language recognition also has been considered in the context of
sequence prediction, with
HMMs~\cite{Ulrich2008,Forster2013,Starner95,Kadir04a},
LSTMs~\cite{Camgoz17,Huang2018VideobasedSL,Ye18,zhou2020spatialtemporal},
and Transformers~\cite{camgoz2020sign} proving to be effective mechanisms
for this task. Recently, convolutional neural networks have emerged as
the dominant approach for appearance modelling~\cite{Camgoz17}, and in
particular, action recognition
models using spatio-temporal convolutions~\cite{Carreira2017} have proven very well-suited for
video-based sign
recognition~\cite{Joze19msasl,Li19wlasl,bsl1k2020}. We adopt the I3D
architecture~\cite{Carreira2017} as a foundational building block in our studies.

\noindent\textbf{Sign language spotting.}
The sign language spotting problem---in which the objective is to find performances of a sign (or sign sequence) in a longer sequence of signing---has been studied with Dynamic Time Warping and skin colour histograms~\cite{viitaniemi14} and with Hierarchical Sequential Patterns~\cite{ong2014}. Different from our work which learns representations from multiple weak supervisory cues, these approaches consider a fully-supervised setting with a single source of supervision and use hand-crafted features to represent signs~\cite{Farhadi07}. 
Our proposed use of a dictionary is also closely tied to \textit{one-shot/few-shot learning}, in which the learner is assumed to have access to only a handful of annotated examples of the target category. One-shot dictionary learning was studied by~\cite{Pfister14} -- different to their approach, we explicitly account for dialect variations in the dictionary (and validate the improvements brought by doing so in Sec.~\ref{sec:experiments}).  Textual descriptions from a dictionary of 250 signs were used to study zero-shot learning by~\cite{Bilge19ZS} -- we instead consider the practical setting in which a handful of video examples are available per-sign (and make this dictionary available). The use of dictionaries to locate signs in subtitled video also shares commonalities with \textit{domain adaptation}, since our method must bridge differences between the dictionary and the target continuous signing distribution. A vast number of techniques have been proposed to tackle distribution shift, including several adversarial feature alignment methods that are specialised for the few-shot setting~\cite{Motiian2017FewShotAD,Zhang2019}.  
In our work, we explore the domain-specific batch normalization (DSBN) method of~\cite{chang2019domain}, finding ultimately that simple batch normalization parameter re-initialization is most effective when jointly training on two domains after pre-training on the bigger domain. The concurrent work of~\cite{li2020transferring} also seeks to align representation of isolated and continuous signs. 
However, our work differs from theirs in several key aspects: (1) rather than assuming access to a large-scale labelled dataset of isolated signs, we consider the setting in which only a handful of dictionary examples may be used to represent a word; (2) we develop a generalised Multiple Instance Learning framework which allows the learning of representations from weakly aligned subtitles whilst exploiting sparse labels and dictionaries (this integrates cues beyond the learning formulation in~\cite{li2020transferring}); (3) we seek to label and improve performance on co-articulated signing (rather than improving recognition performance on isolated signing). Also related to our work, \cite{Pfister14} uses a \say{reservoir} of weakly labelled sign footage to improve the performance of a sign classifier learned from a small number of examples. Different to~\cite{Pfister14}, we propose a multi-instance learning formulation that explicitly accounts for signing variations that are present in the dictionary. 

\noindent\textbf{Sign language datasets.} A number of sign language datasets have been proposed for studying Finnish~\cite{viitaniemi14}, German~\cite{Koller15cslr,signum2008}, American~\cite{asllvid2008,Joze19msasl,Li19wlasl,purdue06} and Chinese~\cite{chai2014devisign,Huang2018VideobasedSL} sign recognition. For British Sign Language (BSL), \cite{schembri2013building} gathered a corpus labelled with sparse, but fine-grained linguistic annotations, and more recently~\cite{bsl1k2020} collected BSL-1K, a large-scale dataset of BSL signs that were obtained using a mouthing-based keyword spotting model. In this work, we contribute~\bsldict{}, a dictionary-style dataset that is complementary to the datasets of~\cite{schembri2013building,bsl1k2020} -- it contains only a handful of instances of each sign, but achieves a comprehensive coverage of the BSL lexicon with a 9K vocabulary (vs a 1K vocabulary in~\cite{bsl1k2020}). As we show in the sequel, this dataset enables a number of sign spotting applications.

\noindent\textbf{Multiple instance learning.} Motivated by the readily available sign language footage that is accompanied by subtitles, a number of methods have been proposed for learning the association between signs and words that occur in the subtitle text \cite{Buehler09,Cooper2009,Pfister14,Chung16b}.  In this work, we adopt the framework of Multiple Instance Learning (MIL)~\cite{dietterich1997solving} to tackle this problem, previously explored by~\cite{Buehler09,pfister2013large}. Our work differs from these works through the incorporation of a dictionary, and a principled mechanism for explicitly handling sign variants, to guide the learning process. Furthermore, we generalise the MIL framework so that it can learn to further exploit sparse labels. We also conduct experiments at significantly greater scale to make use of the full potential of MIL, considering more than two orders of magnitude more weakly supervised data than~\cite{Buehler09,pfister2013large}.

\noindent\textbf{Low-shot action localization.} 
This theme investigates semantic video localization: given one or more query videos the objective is to localize the segment in an untrimmed video that corresponds  semantically to the query video~\cite{Feng_2018_ECCV,Yang_2018_CVPR,Cao_2020_CVPR}.
Semantic matching is too general for the sign-spotting considered in this paper. However, we build on the temporal ordering ideas explored in this theme.

%\mbox{}\vspace{-1.2cm}\\
\section{Learning Sign Spotting Embeddings from Multiple Supervisors} \label{sec:method}

\begin{figure}[t]
    \centering
    \includegraphics[width=0.9\textwidth]{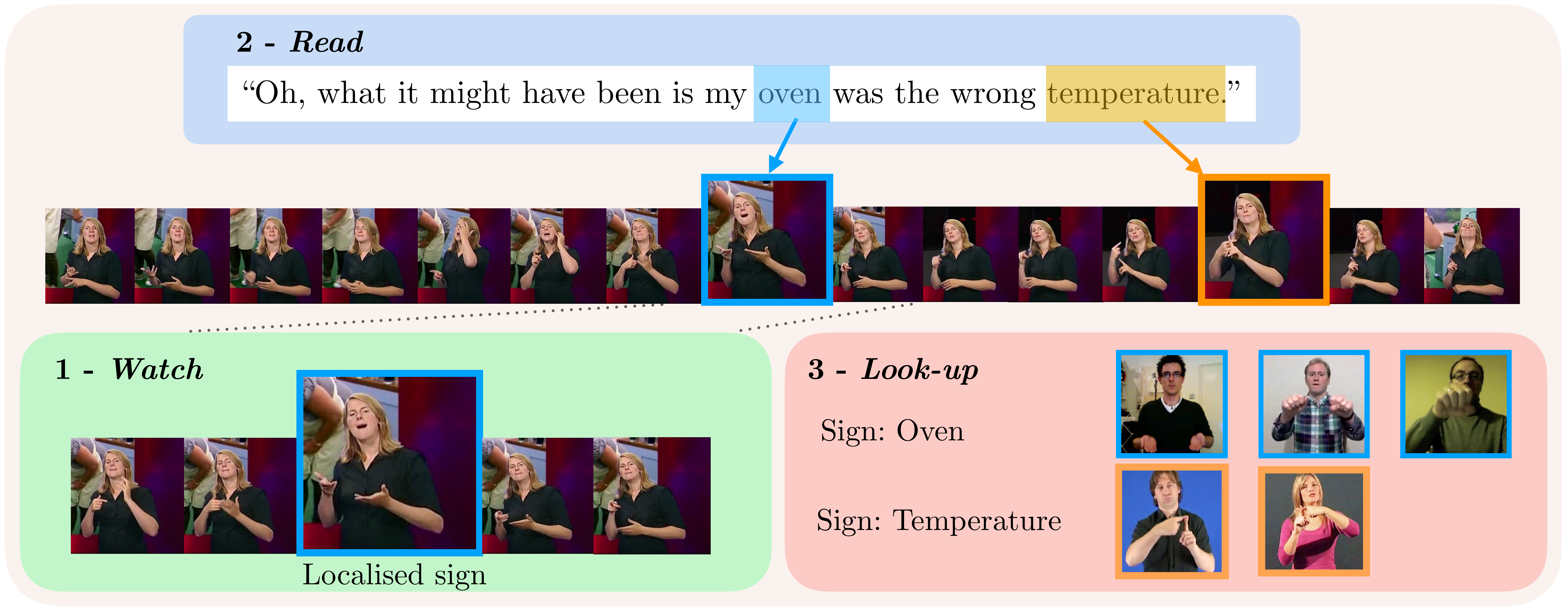}
    %\mbox{}\vspace{-0.2cm}\\
    \caption{\textbf{The proposed \textit{Watch, Read and Lookup} framework} trains sign spotting embeddings with three cues: (1) \textit{watching} videos and learning from sparse annotation in the form of localised signs (lower-left); (2) \textit{reading} subtitles to find candidate signs that may appear in the source footage (top); (3) \textit{looking up} corresponding visual examples in a sign language dictionary and aligning the representation against the embedded source segment (lower-right).}
    %\mbox{}\vspace{-0.8cm}\\
    \label{fig:watch-read-lookup}
\end{figure}

In this section, we describe the task of \textit{sign spotting} and the three forms of supervision we assume access to. Let $\mathcal{X}_{\mathfrak{L}}$ denote the space of RGB video segments containing a frontal-facing individual communicating in sign language $\mathfrak{L}$ and denote by $\mathcal{X}_{\mathfrak{L}}^{\text{single}}$ its restriction to the set of segments containing a single sign. Further, let $\mathcal{T}$ denote the space of subtitle sentences and $\mathcal{V}_{\mathfrak{L}}=~\{1, \dots, V \}$ denote the \textit{vocabulary}---an index set corresponding to an enumeration of written words that are equivalent to signs that can be performed in $\mathfrak{L}$\footnote{Sign language dictionaries provide a word-level or phrase-level correspondence
(between sign language and spoken language) for many signs but no universally accepted \textit{glossing} scheme exists for transcribing languages such as BSL~\cite{sutton-spence_woll_1999}.}.

Our objective, illustrated in Fig.~\ref{fig:kws-task}, is to discover all occurrences of a given keyword in a collection of continuous signing sequences.  To do so, we assume access to: (i) a subtitled collection of videos containing continuous signing, $\mathcal{S} = \{(x_i, s_i ) : i \in \{1, \dots, I\}, x_i \in \mathcal{X}_{\mathfrak{L}}, s_i \in \mathcal{T}\}$; (ii) 
a sparse collection of temporal sub-segments of these videos that have been annotated with their corresponding word, $\mathcal{M} = \{(x_k, v_k) : k \in \{1, \dots, K\}, v_k \in \mathcal{V}_\mathfrak{L}, x_k \in \mathcal{X}_\mathfrak{L}^{\text{single}}, \exists (x_i, s_i) \in \mathcal{S} \, s.t. \, x_k \subseteq x_i \}$; 
(iii) a curated \textit{dictionary} of signing instances $\mathcal{D} = \{(x_j, v_j) : j \in \{1, \dots, J\}, x_j \in \mathcal{X}_{\mathfrak{L}}^{\text{single}}, v_j \in \mathcal{V}_\mathfrak{L}\}$.
%\footnote{\samuel{Describe equivalence relation defined by glosses, or update terminology if we don't use gloss info.}}
To address the sign spotting task, we propose to learn a \textit{data representation} $f: \mathcal{X}_\mathfrak{L} \rightarrow \mathbb{R}^d$ that maps video segments to vectors such that they are \textit{discriminative} for sign spotting and \textit{invariant} to other factors of variation.  Formally, for any labelled pair of video segments $(x, v), (x', v')$ with $x, x' \in \mathcal{X}_\mathfrak{L}$ and $v, v' \in \mathcal{V}_\mathfrak{L}$, we seek a data representation, $f$, that satisfies the constraint  $\delta_{f(x) f(x')} = \delta_{v v'}$, where $\delta$ represents the Kronecker delta. 

%\mbox{}\vspace{-1.2cm}\\
\subsection{Integrating Cues through Multiple Instance Learning }
\label{subsection:mil}

\begin{figure}[t]
  \centering
  \includegraphics[width=0.98\textwidth]{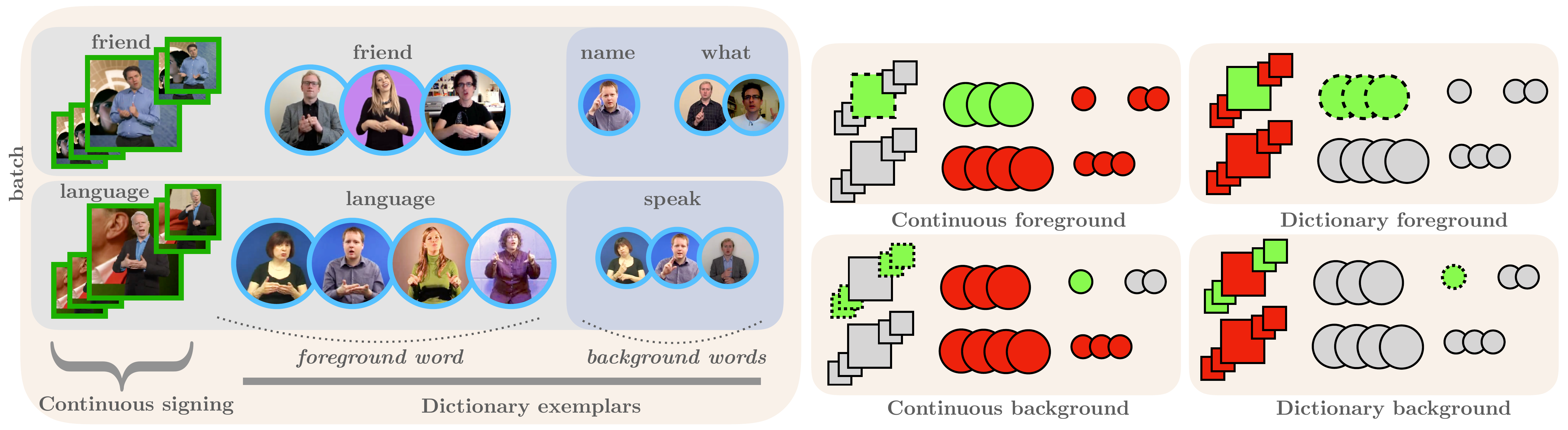}
    %\mbox{}\vspace{-0.5cm}\\
  \caption{\textbf{Batch sampling and positive/negative pairs:} We illustrate the formation of a batch when jointly training on continuous signing video (squares) and dictionaries of isolated signing (circles).
  \textbf{Left:} For each continuous video, we sample the dictionaries corresponding to the labelled word (foreground), as well as %the dictionaries corresponding
  to the rest of the subtitles (background). 
  \textbf{Right:} We construct positive/negative pairs by anchoring at 4 different portions of a batch item: continuous foreground/background and dictionary foreground/background. Positives and negatives (defined across continuous and dictionary domains)
  are green and red, respectively; anchors have a dashed border (see
  \if\sepappendix1{Appendix~C.2}
  \else{Appendix~\ref{app:subsec:math}}
  \fi
  for details).
  % We mark the orange hashed samples as negatives if there is no overlap within subtitles or word labels, otherwise we discard them.
%  \daffy{say that shapes (rectangle, circles) correspond from left to right}
  }
    %\mbox{}\vspace{-0.8cm}\\
  \label{fig:sampling2}
\end{figure}

To learn $f$, we must address several challenges.  First, as noted in Sec.~\ref{sec:intro}, there may be a considerable distribution shift between the dictionary videos of isolated signs in $\mathcal{D}$ and the co-articulated signing videos in $\mathcal{S}$.  Second, sign languages often contain multiple sign variants for a single written word (resulting from regional dialects and synonyms).  Third, since the subtitles in $\mathcal{S}$ are only weakly aligned with the sign sequence, we must learn to associate signs and words from a noisy signal that lacks temporal localisation.  Fourth, the localised annotations provided by $\mathcal{M}$ are sparse, and therefore we must make good use of the remaining segments of subtitled videos in $\mathcal{S}$ if we are to learn an effective representation.

Given full supervision, we could simply adopt a pairwise metric learning approach to align segments from the videos in $\mathcal{S}$ with dictionary videos from $\mathcal{D}$ by requiring that $f$ maps a pair of isolated and co-articulated signing segments to the same point in the embedding space if they correspond to the same sign (\textit{positive} pairs) and apart if they do not (\textit{negative} pairs).
As noted above, in practice we do not have access to positive pairs because: (1) for any annotated segment $(x_k, v_k) \in \mathcal{M}$, we have a set of potential sign variations represented in the dictionary (annotated with the common label $v_k$), rather than a single unique sign; (2) since $\mathcal{S}$ provides only weak supervision, even when a word is mentioned in the subtitles we do not know where it appears in the continuous signing sequence (if it appears at all). 
These ambiguities motivate a Multiple Instance Learning~\cite{dietterich1997solving} (MIL) objective. Rather than forming positive and negative pairs, we instead form positive \textit{bags} of pairs, $\mathcal{P}^{\text{bags}}$, in which we expect at least one pairing between a segment from a video in $\mathcal{S}$ and a dictionary video from $\mathcal{D}$ to contain the same sign, and negative bags of pairs, $\mathcal{N}^{\text{bags}}$, in which we expect no (video segment, dictionary video) pair to contain the same sign.  To incorporate the available sources of supervision into this formulation, we consider two categories of positive and negative bag formations, described next (due to space constraints, a formal mathematical description of the positive and negative bags described below is deferred to
\if\sepappendix1{Appendix~C.2).}
\else{Appendix~\ref{app:subsec:math}).}
\fi
% to the supp.~mat.).

\noindent \textbf{Watch and Lookup: using sparse annotations and dictionaries}. Here, we describe a baseline where we assume no subtitles are available. To learn $f$ from $\mathcal{M}$ and $\mathcal{D}$, we define each positive bag as the set of possible pairs between a \textit{labelled} (\textit{foreground}) temporal segment of a continuous video from $\mathcal{M}$ and the examples of the corresponding sign in the dictionary (green regions in
\if\sepappendix1{Fig~A.2).}
\else{Fig~\ref{app:fig:sampling_milnce}).}
\fi
The key assumption here is that each labelled sign segment  from $\mathcal{M}$ matches \textit{at least one} sign variation in the dictionary.
Negative bags are constructed by (i) anchoring on a continuous
foreground segment and selecting dictionary examples
corresponding to different words
from other batch items; (ii) anchoring on
a dictionary foreground set and
selecting continuous foreground segments from other batch items
(red regions in
\if\sepappendix1{Fig~A.2).}
\else{Fig~\ref{app:fig:sampling_milnce}).}
\fi
To maximize the number of negatives
within one minibatch, we sample a different word per batch item.

% Negative bags are constructed by (i) anchoring on continuous foreground segments and selecting dictionary examples corresponding to different words; (ii) anchoring on the dictionary foreground set for a given word and selecting pairs among segments that do not overlap with the labelled segment (visualised as the red regions in Fig.~\ref{fig:sampling2},  right-top).

\noindent \textbf{Watch, Read and Lookup: using sparse annotations, subtitles and dictionaries}. Using just the labelled sign segments from $\mathcal{M}$ to construct bags has a significant limitation: $f$ is not encouraged to represent signs beyond the initial vocabulary represented in $\mathcal{M}$.  We therefore look at the subtitles (which contain words beyond $\mathcal{M}$) to construct additional bags. We determine more positive bags between the set of \textit{unlabelled (background)} segments in the continuous footage and the set of dictionaries corresponding to the background words in the subtitle (green regions in Fig.~\ref{fig:sampling2},  right-bottom). Negatives (red regions in Fig.~\ref{fig:sampling2}) are formed as the complements to these sets by (i) pairing continuous background segments with dictionary samples that can be excluded as matches (through subtitles) and (ii) pairing background dictionary entries with the foreground continuous segment.
In both cases, we also define negatives from other batch items
by selecting pairs where the word(s) have no overlap, e.g.,
in~Fig.~\ref{fig:sampling2}, the dictionary examples for the background word `speak' from the second batch item are negatives
for the background continuous segments from the first batch item, corresponding to the unlabelled words `name' and `what' in the subtitle.
% , because the two subtitles do not have any words in common.

To assess the similarity of two embedded video segments, we employ a similarity function $\psi: \mathbb{R}^d \times \mathbb{R}^d \rightarrow \mathbb{R}$ whose value increases as its arguments become more similar (in this work, we use cosine similarity). For notational convenience below, we write $\psi_{ij}$ as shorthand for $\psi(f(x_i), f(x_j))$. 
To learn $f$, we consider a generalization of the InfoNCE loss~\cite{oord2018representation,wu2018unsupervised} (a non-parametric softmax loss formulation of Noise Contrastive Estimation~\cite{gutmann2010noise}) recently proposed by~\cite{miech2020end}:

\begin{align}
    \mathcal{L}_{\text{MIL-NCE}} = - \mathbb{E}_i \Bigg[ \log \frac{\sum_{(j,k) \in \mathcal{P}(i)} \exp(\psi_{jk}/ \tau)}{\sum_{(j,k) \in \mathcal{P}(i)} \exp(\psi_{jk}/ \tau) +  \sum_{(l,m) \in \mathcal{N}(i)} \exp(\psi_{lm}/ \tau)} \Bigg], \label{eqn:mil-nce}
\end{align}

where $\mathcal{P}(i) \in \mathcal{P}^{\text{bags}}$, $\mathcal{N}(i) \in \mathcal{N}^{\text{bags}}$, $\tau$, often referred to as the \textit{temperature}, is set as a hyperparameter (we explore the effect of its value in Sec.~\ref{sec:experiments}).

\subsection{Implementation details }\label{subsection:implementation}

In this section, we provide details for the learning framework covering the embedding architecture, sampling protocol and optimization procedure.

\noindent{\textbf{Embedding architecture.}} The architecture comprises an I3D spatio-temporal trunk network~\cite{Carreira2017} to which we attach an MLP consisting of three linear layers separated by leaky ReLU activations (with negative slope 0.2) and a skip connection. The trunk network takes as input 16 frames from a $224\times224$ resolution video clip and produces $1024$-dimensional embeddings which are then projected to 256-dimensional sign spotting embeddings by the MLP.
More details about the embedding architecture can be found in
\if\sepappendix1{Appendix~C.1.}
\else{Appendix~\ref{app:subsec:arch}.}
\fi
% the supplementary material.

\noindent{\textbf{Joint pretraining.}} The I3D trunk parameters are initialised by pretraining for sign classification jointly over the sparse annotations $\mathcal{M}$ of a continuous signing dataset (\bslonek{}~\cite{bsl1k2020}) and examples from a sign dictionary dataset (\bsldict) which fall within their common vocabulary.
Since we find that dictionary videos of isolated signs tend to be performed more slowly, we uniformly sample 16 frames from each dictionary video with a random shift and random frame rate $n$ times, where $n$ is proportional to the length of the video, and pass these clips through the I3D trunk then average the resulting vectors before they are processed by the MLP to produce the final dictionary embeddings. We find that this form of random sampling performs better than sampling 16 consecutive frames from the isolated signing videos (see
\if\sepappendix1{Appendix~C.1}
\else{Appendix~\ref{app:subsec:arch}}
\fi
% supplementary material
for more details).
During pretraining, minibatches of size 4 are used; and colour, scale and horizontal flip augmentations are applied to the input video, following the procedure described in~\cite{bsl1k2020}.
The trunk parameters are then frozen and the MLP outputs are used as embeddings. Both datasets are described in detail in Sec.~\ref{subsection:datasets}.

\noindent{\textbf{Minibatch sampling.}} To train the MLP given the pretrained I3D features, we sample data by first iterating over the set of labelled segments comprising the sparse annotations, $\mathcal{M}$, that accompany the dataset of continuous, subtitled sampling to form minibatches.
For each continuous video, we sample 16 consecutive frames around the annotated timestamp (more precisely a random offset within 20 frames before, 5 frames after, following the timing study in~\cite{bsl1k2020}). We randomly sample 10 additional 16-frame clips from this video outside of the labelled window, i.e., continuous background segments.
For each subtitled sequence, we sample the dictionary entries for all subtitle words that appear in $\mathcal{V}_{\mathfrak{L}}$ (see Fig.~\ref{fig:sampling2} for a sample batch formation).

Our minibatch comprises 128 sequences of continuous signing and their corresponding dictionary entries (we investigate the impact of batch size in Sec.~\ref{subsection:ablations}). The embeddings are then trained by minimising the loss defined in Eqn.(\ref{eqn:mil-nce}) in conjunction with 
positive bags, $\mathcal{P}_{\text{}}^{\text{bags}}$, and negative bags, $\mathcal{N}_{\text{}}^{\text{bags}}$, which are constructed on-the-fly for each minibatch (see Fig.~\ref{fig:sampling2}).

\noindent{\textbf{Optimization.}} We use a SGD optimizer with an initial learning rate of $10^{-2}$ to train the embedding architecture. The learning rate is decayed twice by a factor of 10 (at epoch 40 and 45). We train all models, including baselines and ablation studies, for 50 epochs at which point we find that learning has always converged.

\noindent{\textbf{Test time.}} To perform spotting, we obtain the embeddings learned with the MLP.
For the dictionary, we have a single embedding averaged over the video.
Continuous video embeddings are obtained with sliding window (stride 1) on the entire sequence.
We calculate the cosine similarity score between the continuous signing sequence embeddings and the embedding for a given dictionary video. We determine the location with the maximum similarity as the location of the queried sign. We maintain embedding sets of all variants of dictionary videos for a given word and choose the best match as the one with the highest similarity.

\section{Experiments} \label{sec:experiments}

In this section, we first present the datasets used in this work (including the contributed \bsldict{} dataset) in Sec.~\ref{subsection:datasets}, followed by the evaluation protocol in Sec.~\ref{subsec:evaluation}.
We illustrate the benefits of the \textit{Watch, Read and Lookup} learning framework for sign spotting
against several baselines 
with a comprehensive ablation study that validates our design choices (Sec.~\ref{subsection:ablations}).
Finally, we investigate three applications of our method in Sec.~\ref{subsection:applications}, showing that it can be used to (i)~not only spot signs, but also identify the specific sign variant that was used, (ii)~label 
%\daffy{better say novel - as in out-of-vocabulary? make more clear}
sign instances in continuous signing footage given the associated subtitles, and (iii)~discover \say{faux amis} between different sign languages.  

\begin{table}[t]
 \centering
 \setlength{\tabcolsep}{8pt}
    \resizebox{0.5\linewidth}{!}{
\begin{tabular}{c c c c }
\toprule
Dataset & \#Videos & Vocab. & \#Signers \\ 
\midrule
 \bslonek\cite{bsl1k2020} &  273K & 1,064 & 40\\
\midrule
\bsldict & 14,210 & 9,283 & 148 \\
\bottomrule
\end{tabular}
}
    \caption{
        \textbf{Datasets:} We provide (i) the number of individual sign videos, (ii) the vocabulary size of the annotated signs, and (iii) the number of signers
        for \bslonek{} and \bsldict. \bslonek{}
        is large in the number of annotated signs whereas \bsldict{} is large in the vocabulary size. Note that
        we use a different partition of \bslonek{} with longer
        sequences around the annotations as described
        in Sec.~\ref{subsection:datasets}.
    }
    \label{tab:datasets}
\end{table}

\subsection{Datasets}\label{subsection:datasets}
Although our method is conceptually applicable to a number of sign languages, in this work we focus primarily on BSL, the sign language of British deaf communities. We
%selected BSL due to the recent development of
use \bslonek~\cite{bsl1k2020}, a large-scale, subtitled and sparsely annotated dataset of more than 1000 hours of continuous signing which offers an ideal setting in which to evaluate the effectiveness of the \textit{Watch, Read and Lookup} sign spotting framework.  To provide dictionary data for the \textit{lookup} component of our approach, we also contribute~\bsldict{}, a diverse visual
dictionary of signs. These two datasets are summarised in Table \ref{tab:datasets} and described in more detail below.

\noindent\textbf{\bslonek}~\cite{bsl1k2020} comprises a vocabulary of 1,064 signs which are sparsely annotated over 1,000 hours of video of continuous sign language. The videos are accompanied by subtitles. The dataset consists of 273K localised sign annotations, automatically generated from sign-language-interpreted BBC television broadcasts, by leveraging weakly aligned subtitles and applying keyword spotting to signer \textit{mouthings}.
Please refer to~\cite{bsl1k2020} for more details on the automatic annotation pipeline. In this work,
we process this data to extract long videos with subtitles. In particular, we pad +/-2 seconds around the subtitle timestamps and we add the corresponding video to our training set if there is a sparse annotation word falling within this time window, assuming that the signing is reasonably well-aligned with its subtitles in these cases. We further consider only the videos whose subtitle duration is longer than 2 seconds. For testing, we use the automatic test set (corresponding to mouthing locations with confidences above 0.9). Thus we obtain 78K training and 3K test videos, each of which has a subtitle of 8 words on average and 1 sparse mouthing annotation. \\

\noindent\textbf{\bsldict}. BSL dictionary videos are collected from a BSL sign aggregation platform \url{signbsl.com}~\cite{signbslcom}, giving us a total of 14,210 video clips for a vocabulary of 9,283 signs. Each sign is typically performed several times by different signers, often in different ways. The dictionary videos are downloaded from 28 known website sources and each source has at least 1 signer. We used face embeddings computed with SENet-50~\cite{hu2019squeeze} (trained on VGGFace2~\cite{Cao18}) to cluster signer identities and manually verified that there are a total of 148 different signers. The dictionary videos are of isolated signs (as opposed to co-articulated in \bslonek): this means (i) the start and end of the video clips usually consist of a still signer pausing, and (ii) the sign is performed at a much slower rate for clarity. We first trim the sign dictionary videos, using body keypoints estimated with OpenPose~\cite{cao2018openpose} which indicate the start and end of wrist motion, to discard frames where the signer is still. With this process, the average number of frames per video drops from 78 to 56 (still significantly larger than co-articulated signs).
To the best of our knowledge, \bsldict{} is the first curated, BSL sign dictionary dataset for computer vision research, which will be made available. For the experiments in which \bsldict{} is filtered to the 1,064 vocabulary of \bslonek{} (see below), we have a total of 2,992 videos. Within this subset, each sign has between 1 and 10 examples (average of 3).

\subsection{Evaluation Protocols}\label{subsec:evaluation}
\noindent{\textbf{Protocols.}} We define two settings: (i) training with the entire 1064 vocabulary of annotations in \bslonek{}; and (ii) training on a subset with 800 signs. The latter is needed to assess the performance on novel signs, for which we do not have access to co-articulated labels at training. We thus use the remaining 264 words for testing. This test set is therefore common to both training settings, it is either `seen' or `unseen' at training.
However, we do not limit the vocabulary of the dictionary as a practical assumption,
for which we show benefits.

\noindent{\textbf{Metrics.}}
The performance is evaluated based on ranking metrics.
For every sign $s_i$ in the test vocabulary, we first select the \bslonek{} test set clips which have a mouthing annotation of $s_i$ and then record the percentage of dictionary clips of $s_i$ that appear in the first 5 retrieved results, this is the ‘Recall at 5’ (R@5). This is motivated by the fact that different English words can correspond to the same sign, and vice versa. We also report mean average precision (mAP). For each video pair, the match is considered correct if (i) the dictionary clip corresponds to $s_i$  and the \bslonek{} video clip has a mouthing annotation of $s_i$, and (ii) if the predicted location of the sign in the \bslonek{} video clip, i.e.\ the time frame where the maximum similarity occurs, lies within certain frames around the ground truth mouthing timing. In particular, we determine the correct interval to be defined between 20 frames before and 5 frames after the labelled time (based on the study in~\cite{bsl1k2020}).
Finally, because \bslonek{} test is class-unbalanced, we report performances averaged over the test classes.

\begin{table}[t]
    \setlength{\tabcolsep}{8pt}
    \centering
    \resizebox{0.9\linewidth}{!}{
        \begin{tabular}{llcc|cc}
            \toprule
            & & \multicolumn{2}{c|}{Train (1064)} & \multicolumn{2}{c}{Train (800)} \\
            \midrule
            & & \multicolumn{2}{c|}{Seen (264)} & \multicolumn{2}{c}{Unseen (264)} \\
            Embedding arch. & Supervision & mAP & R@5 & mAP & R@5 \\
            \midrule
            \baselineDict{} & Classification & 2.68 & 3.57 & 1.21 & 1.29 \\
            \baselineBbc{} \cite{bsl1k2020} & Classification & 13.09 & 17.25 & 6.74 & 8.94 \\
            \baselineBbcDict{} & Classification & 19.81 &  25.57 & 4.81 & 6.89 \\ 
            \baselineBbcDict{}+MLP & Classification & 36.75 & 40.15 & 10.28 & 14.19 \\
            \midrule
            \baselineBbcDict{}+MLP & InfoNCE & 42.52 & 53.54 & 10.88 & 14.23 \\
            \baselineBbcDict{}+MLP & Watch-Lookup & 43.65 & 53.03 & 11.05 & 14.62 \\
            \baselineBbcDict{}+MLP & Watch-Read-Lookup & \textbf{48.11} & \textbf{58.71} & \textbf{13.69} & \textbf{17.79} \\
            \bottomrule
        \end{tabular}
    }
    \caption{\textbf{The effect of the loss formulation:} Embeddings learned with the classification loss are suboptimal since they are not trained for matching the two domains. Contrastive-based loss formulations (NCE) significantly improve, particularly when we adopt the multiple-instance variant introduced as our Watch-Read-Lookup framework of multiple supervisory signals.
    }
    %\mbox{}\vspace{-0.8cm}\\
    \label{tab:loss}
\end{table}

\subsection{Ablation Study}\label{subsection:ablations}
In this section, we evaluate different components of our approach.
We first compare our contrastive learning approach with classification baselines.
Then, we investigate the effect of our multiple-instance loss formulation.
We provide ablations for the hyperparameters, such as the batch size and the temperature, and report performance on a sign spotting benchmark.

\noindent\textbf{I3D baselines.}
We start by evaluating baseline I3D models trained with classification
on the task of spotting, using the embeddings before the classification layer.
We have three variants in Tab.~\ref{tab:loss}: (i) \baselineBbc{} provided by \cite{bsl1k2020}
which is trained only on the \bslonek{} dataset, and we also train (ii) \baselineDict{}
and (iii) \baselineBbcDict{}. Training only on \bsldict{} (\baselineDict{})
performs significantly worse due to the few examples available per class and the domain gap that must be bridged to spot co-articulated signs, suggesting that dictionary samples alone do not suffice to solve the task. We observe
improvements with fine-tuning
\baselineBbc{} jointly on the two datasets (\baselineBbcDict{}),
which becomes our base feature extractor for the remaining experiments
to train a shallow MLP.

\noindent\textbf{Loss formulation.}
We first train the MLP parameters on top of the frozen I3D trunk with classification to establish a baseline in a comparable setup.
Note that, this shallow architecture can be trained with larger batches than I3D.
Next, we investigate variants of our loss to learn a joint sign embedding between \bslonek{} and \bsldict{} video domains: (i) standard single-instance InfoNCE~\cite{oord2018representation,wu2018unsupervised} loss which pairs each \bslonek{} video clip with \textit{one} positive \bsldict{} clip of the same sign, (ii) Watch-Lookup which considers
multiple positive dictionary candidates, but does not consider subtitles (therefore limited to the annotated video clips). Table~\ref{tab:loss}
summarizes the results. Our Watch-Read-Lookup formulation which
effectively combines multiple sources of supervision
in a multiple-instance framework outperforms the other baselines
in both \textit{seen} and \textit{unseen} protocols.

\noindent\textbf{Extending the vocabulary.} The results presented so far
were using the same vocabulary for both continuous and dictionary datasets. In reality,
one can assume access to the entire vocabulary in the dictionary, but obtaining
annotations for the continuous videos is prohibitive.
Table~\ref{tab:vocab}
investigates removing the vocabulary limit on the dictionary side, but keeping
the continuous annotations vocabulary at 800 signs. We show that using the full 9k vocabulary from
\bsldict{} significantly improves the results on the unseen setting.

\begin{table}[t]
    \setlength{\tabcolsep}{8pt}
    \centering
    \resizebox{0.7\linewidth}{!}{
        \begin{tabular}{llcc}
            \toprule
            Supervision & Dictionary Vocab & mAP & R@5 \\
            \midrule
            Watch-Read-Lookup & 800 training vocab & 13.69 & 17.79 \\ 
            Watch-Read-Lookup & 9k full vocab & \textbf{15.39} & \textbf{20.87} \\ 
            \bottomrule
        \end{tabular}
    }
    \caption{\textbf{Extending the dictionary vocabulary:} We show the benefits of sampling dictionary videos outside of the sparse annotations, using subtitles. Extending the lookup to the dictionary from the subtitles to the full vocabulary of \bsldict{} brings significant improvements for novel signs (the training uses sparse annotations for the 800 words, and the remaining 264 for test).
    }
    %\mbox{}\vspace{-0.8cm}\\
    \label{tab:vocab}
\end{table}

\setlength{\tabcolsep}{1pt}
\begin{figure}[t]
    \centering
    \begin{tabular}{llll}
    (a) & \raisebox{-.5\height}{\includegraphics[width=0.45\textwidth]{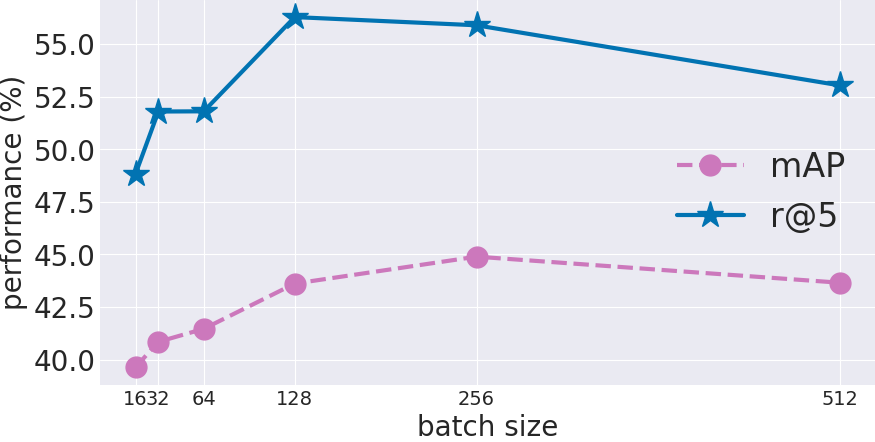}} & (b) & \raisebox{-.5\height}{\includegraphics[width=0.45\textwidth]{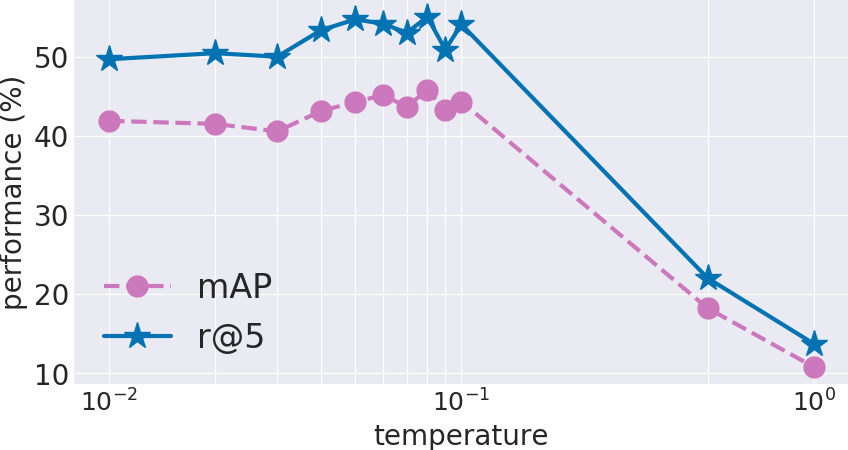}} %\\
    \end{tabular}
    \caption{The effect of (a) the \textbf{batch size} that determines the number of negatives across sign classes and (b) the \textbf{temperature} hyper-parameter for the MIL-NCE loss in Watch-Lookup against mAP and R@5 (trained on the full 1064 vocab.)}
    %\mbox{}\vspace{-0.6cm}\\
    \label{fig:hyperparams}
\end{figure}

\noindent\textbf{Batch size.} Next, we investigate the effect of increasing the number of negative pairs by increasing the batch size when training with Watch-Lookup on 1064 categories. We observe in Figure~\ref{fig:hyperparams}(a) an improvement in performance with greater numbers of negatives before saturating.
Our final Watch-Read-Lookup model has high memory requirements, for which we use 128 batch size. Note that the effective size of the batch with our sampling is larger
due to sampling extra video clips corresponding to subtitles.

\noindent\textbf{Temperature.} Finally, we analyze the impact of the temperature hyperparameter $\tau$ on the performance of Watch-Lookup. We observe a major decrease in performance when $\tau$ approaches 1. We choose $\tau = 0.07$ used in~\cite{wu2018unsupervised,he2020momentum} for all other experiments. Additional ablations are provided in
\if\sepappendix1{Appendix~B.}
\else{Appendix~\ref{app:sec:additionalexp}.}
\fi
% the supplementary material.

\noindent\textbf{\bslonek{} Sign spotting benchmark.} Although our learning framework primarily targets good performance on unseen continuous signs, it can also be naively applied to the (closed-vocabulary) sign spotting benchmark proposed by~\cite{bsl1k2020}. We evaluate the performance of our Watch-Read-Lookup model and achieve a score of 0.170 mAP, outperforming the previous state-of-the-art performance of 0.160 mAP~\cite{bsl1k2020}.

\subsection{Applications}\label{subsection:applications}
In this section, we investigate three applications of our sign spotting method.

\begin{figure}[t]
  \centering
  \includegraphics[width=0.97\textwidth]{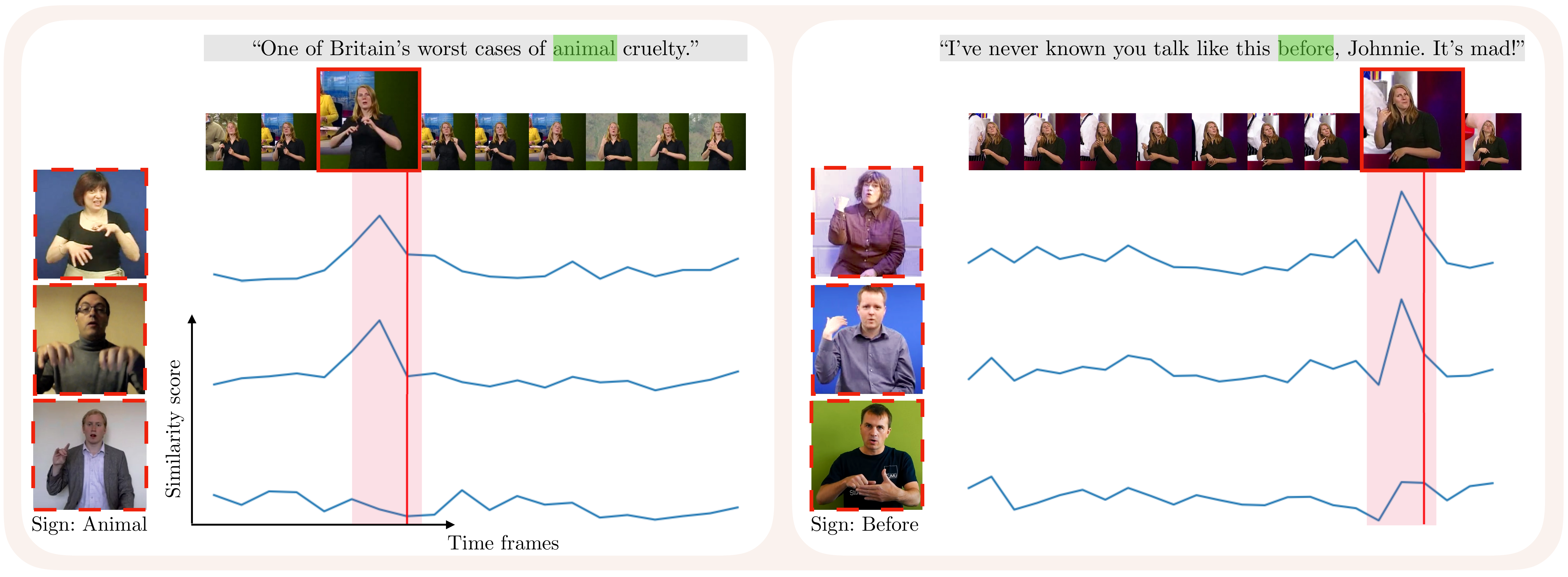}
  \caption{\textbf{Sign variant identification:} We plot the similarity scores between \bslonek{} test clips and \bsldict{} variants of the sign ``animal'' (left) and ``before'' (right) over time. The labelled mouthing times are shown by red vertical lines and the sign proposal regions are shaded. %, both of which are obtained from \bslonek{}.
  A high similarity occurs for the first two rows, where the \bsldict{} examples match the variant used in \bslonek{}. }
  \mbox{}\vspace{-0.8cm}\\
  \label{fig:variants}
\end{figure}

\noindent\textbf{Sign variant identification.} We show the ability of our model to spot
% not only the sign corresponding to a word, but more
specifically which variant of the sign was used. In Fig.~\ref{fig:variants}, we observe high similarity scores when the variant of the sign matches in both \bslonek{} and \bsldict{} videos.  Identifying such sign variations allows a better understanding of regional differences and can potentially help standardisation efforts of BSL.
 %\\

\begin{figure}[t]
  \centering
  \includegraphics[width=0.98\textwidth]{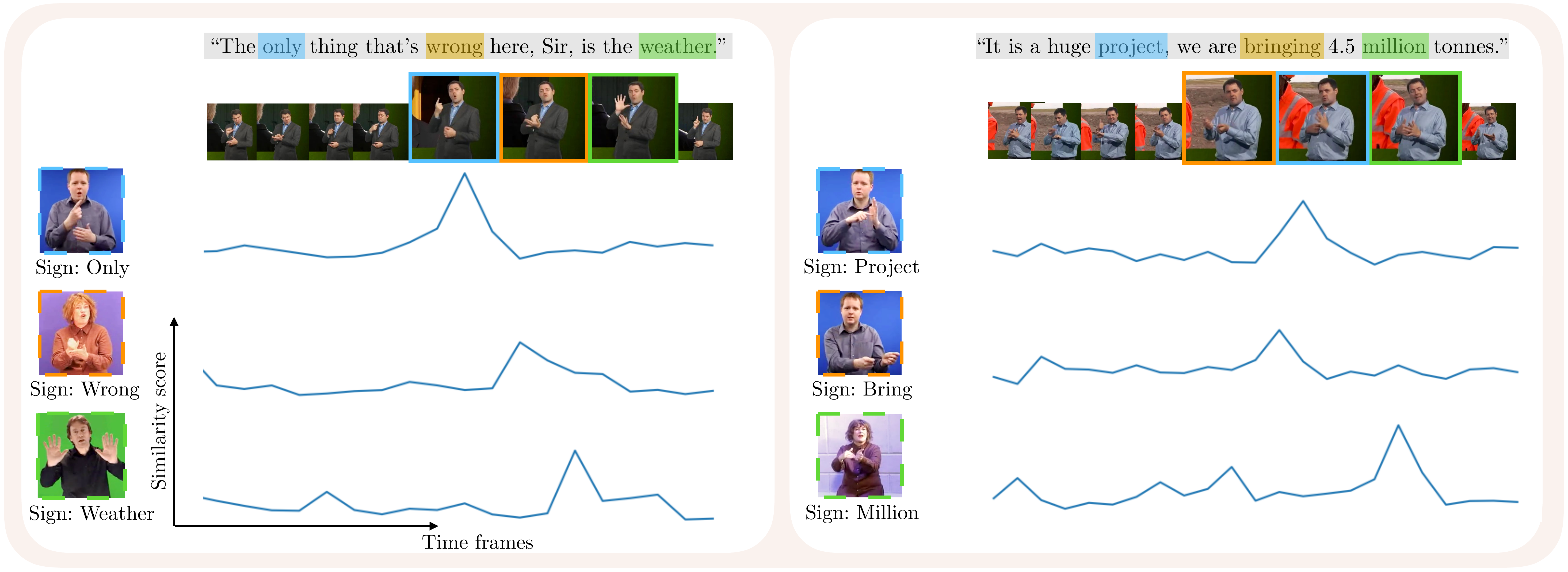}
  \caption{\textbf{Densification:} We plot the similarity scores between \bslonek{} test clips and \bsldict{} examples over time, by querying only the words in the subtitle. The predicted locations of the signs correspond to the peak similarity scores.}
  %\mbox{}\vspace{-0.6cm}\\
  \label{fig:densification}
\end{figure}

\noindent\textbf{Dense annotations.} We demonstrate the potential of our model to obtain dense annotations
on continuous sign language video data. Sign spotting through the use of sign dictionaries is not limited to mouthings as in~\cite{bsl1k2020} and therefore is of
great importance to scale up datasets for learning more robust sign language models.
In Fig.~\ref{fig:densification}, we show qualitative examples
of localising multiple signs in a given sentence in \bslonek{}, where we only query
the words that occur in the subtitles, reducing the search space. In fact, if we assume the word to be known, we obtain {83.08\%} sign localisation accuracy on \bslonek{} with our best model. This is defined as the number of times the maximum similarity occurs within -20/+5 frames of the end label time provided by~\cite{bsl1k2020}. % \\

\begin{figure}[t]
  \centering
  \includegraphics[width=0.98\textwidth]{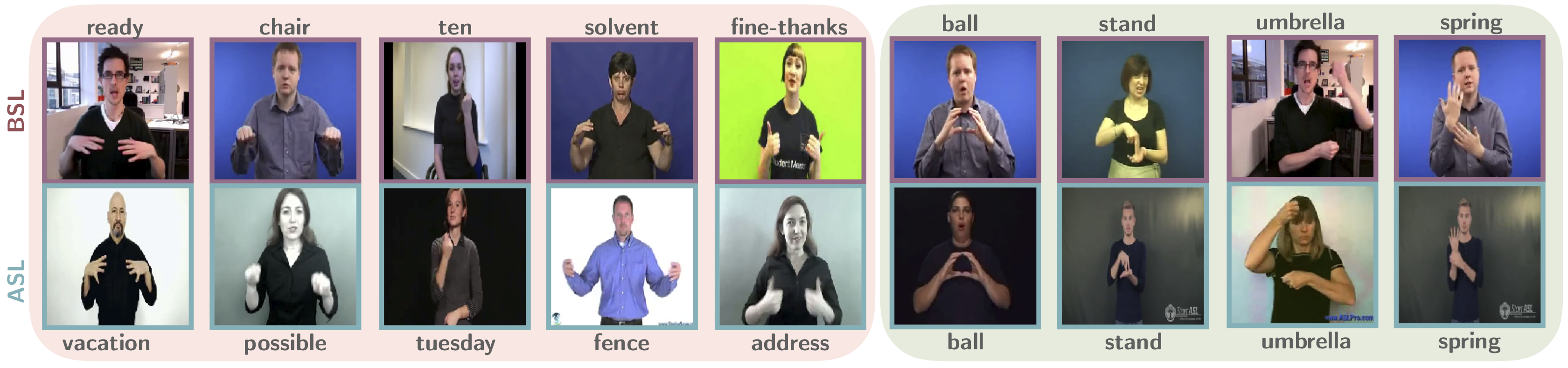}
  \caption{\textbf{``Faux amis'' in BSL/ASL:}  Same/similar manual features for different English words (left), as well as for the same English words (right), are identified between \bsldict{} and WLASL isolated sign language datasets.}
  \mbox{}\vspace{-0.8cm}\\
  \label{fig:fauxamis}
\end{figure}

\noindent\textbf{``Faux Amis''.} There are works investigating lexical similarities between sign languages manually~\cite{SignumMcKee2000,Aldersson2007}. We show qualitatively the potential of our model to discover similarities, as well as ``faux-amis" between different sign languages, in particular between British (BSL) and American (ASL) Sign Languages. We retrieve nearest neighbors according to visual embedding similarities between \bsldict{} which has a 9K vocabulary and WLASL~\cite{Li19wlasl}, an ASL isolated sign language dataset, with a 2K vocabulary. We provide some examples in Fig.~\ref{fig:fauxamis}.

\section{Conclusions} \label{sec:conclusions}
We have presented an approach to spot signs
in continuous sign language videos using visual sign dictionary videos, and have shown the benefits of leveraging multiple supervisory signals
available in a realistic setting: (i) sparse annotations in continuous signing,
(ii) accompanied with subtitles, and (iii) a few dictionary samples per word
from a large vocabulary. We employ multiple-instance contrastive learning to incorporate
these signals into a unified framework. Our analysis
suggests the potential of sign spotting in several applications, which we think
will help in scaling up the automatic annotation of sign language datasets. %\\

\noindent\textbf{Acknowledgements.}
This work was supported by EPSRC grant ExTol. The authors would to like thank A.~Sophia Koepke, Andrew Brown, Necati Cihan Camg\"oz, and Bencie Woll for their help. S.A would like to acknowledge the generous support of S. Carlson in enabling his contribution, and his son David, who bravely waited until after the submission deadline to enter this world.

\bibliographystyle{splncs}
\bibliography{shortstrings,vgg_local,references}

\begin{thebibliography}{10}

\bibitem{sutton-spence_woll_1999}
Sutton-Spence, R., Woll, B.:
\newblock The Linguistics of {B}ritish Sign Language: An Introduction.
\newblock Cambridge University Press (1999)

\bibitem{coucke2019efficient}
Coucke, A., Chlieh, M., Gisselbrecht, T., Leroy, D., Poumeyrol, M., Lavril, T.:
\newblock Efficient keyword spotting using dilated convolutions and gating.
\newblock In: ICASSP. (2019)

\bibitem{veniat2019stochastic}
V{\'e}niat, T., Schwander, O., Denoyer, L.:
\newblock Stochastic adaptive neural architecture search for keyword spotting.
\newblock In: ICASSP. (2019)

\bibitem{Momeni20}
Momeni, L., Afouras, T., Stafylakis, T., Albanie, S., Zisserman, A.:
\newblock Seeing wake words: Audio-visual keyword spotting.
\newblock In: BMVC. (2020)

\bibitem{stafylakis2018zero}
Stafylakis, T., Tzimiropoulos, G.:
\newblock Zero-shot keyword spotting for visual speech recognition in-the-wild.
\newblock In: ECCV. (2018)

\bibitem{chung2017lip}
Chung, J.S., Senior, A., Vinyals, O., Zisserman, A.:
\newblock Lip reading sentences in the wild.
\newblock In: CVPR. (2017)

\bibitem{afouras2018lrs3}
Afouras, T., Chung, J.S., Zisserman, A.:
\newblock {LRS3-TED}: a large-scale dataset for visual speech recognition.
\newblock arXiv preprint arXiv:1809.00496 (2018)

\bibitem{bsl1k2020}
Albanie, S., Varol, G., Momeni, L., Afouras, T., Chung, J.S., Fox, N.,
  Zisserman, A.:
\newblock {BSL-1K}: Scaling up co-articulated sign recognition using mouthing
  cues.
\newblock In: ECCV. (2020)

\bibitem{bslcorpus17}
Schembri, A., Fenlon, J., Rentelis, R., Cormier, K.:
\newblock {British Sign Language Corpus Project: A corpus of digital video data
  and annotations of British Sign Language 2008-2017 (Third Edition)} (2017)

\bibitem{gutmann2010noise}
Gutmann, M., Hyv{\"a}rinen, A.:
\newblock Noise-contrastive estimation: A new estimation principle for
  unnormalized statistical models.
\newblock In: Proceedings of the Thirteenth International Conference on
  Artificial Intelligence and Statistics. (2010)  297--304

\bibitem{Kadir04a}
Kadir, T., Bowden, R., Ong, E.J., Zisserman, A.:
\newblock Minimal training, large lexicon, unconstrained sign language
  recognition.
\newblock In: Proc. BMVC. (2004)

\bibitem{Tamura88}
Tamura, S., Kawasaki, S.:
\newblock Recognition of sign language motion images.
\newblock Pattern Recognition \textbf{21} (1988)  343 -- 353

\bibitem{Starner95}
Starner, T.:
\newblock Visual recognition of american sign language using hidden markov
  models.
\newblock Master's thesis, Massachusetts Institute of Technology (1995)

\bibitem{Fillbrandt2003}
{Fillbrandt}, H., {Akyol}, S., {Kraiss}, K..:
\newblock Extraction of {3D} hand shape and posture from image sequences for
  sign language recognition.
\newblock In: IEEE International SOI Conference. (2003)

\bibitem{Buehler09}
Buehler, P., Everingham, M., Zisserman, A.:
\newblock Learning sign language by watching {TV} (using weakly aligned
  subtitles).
\newblock In: Proc. CVPR. (2009)

\bibitem{Cooper2011}
{Cooper}, H., {Pugeault}, N., {Bowden}, R.:
\newblock Reading the signs: A video based sign dictionary.
\newblock In: ICCVW. (2011)

\bibitem{Ong2012}
{Ong}, E., {Cooper}, H., {Pugeault}, N., {Bowden}, R.:
\newblock Sign language recognition using sequential pattern trees.
\newblock In: CVPR. (2012)

\bibitem{Pfister14}
Pfister, T., Charles, J., Zisserman, A.:
\newblock Domain-adaptive discriminative one-shot learning of gestures.
\newblock In: Proc. ECCV. (2014)

\bibitem{Ulrich2008}
Agris, U., Zieren, J., Canzler, U., Bauer, B., Kraiss, K.F.:
\newblock Recent developments in visual sign language recognition.
\newblock Universal Access in the Information Society \textbf{6} (2008)
  323--362

\bibitem{Forster2013}
Forster, J., Oberd{\"o}rfer, C., Koller, O., Ney, H.:
\newblock Modality combination techniques for continuous sign language
  recognition.
\newblock In: Pattern Recognition and Image Analysis. (2013)

\bibitem{Camgoz17}
Camgoz, N.C., Hadfield, S., Koller, O., Bowden, R.:
\newblock {SubUNets}: {E}nd-to-end hand shape and continuous sign language
  recognition.
\newblock In: ICCV. (2017)

\bibitem{Huang2018VideobasedSL}
Huang, J., Zhou, W., Zhang, Q., Li, H., Li, W.:
\newblock Video-based sign language recognition without temporal segmentation.
\newblock In: AAAI. (2018)

\bibitem{Ye18}
{Ye}, Y., {Tian}, Y., {Huenerfauth}, M., {Liu}, J.:
\newblock Recognizing american sign language gestures from within continuous
  videos.
\newblock In: CVPRW. (2018)

\bibitem{zhou2020spatialtemporal}
Zhou, H., Zhou, W., Zhou, Y., Li, H.:
\newblock Spatial-temporal multi-cue network for continuous sign language
  recognition.
\newblock CoRR \textbf{abs/2002.03187} (2020)

\bibitem{camgoz2020sign}
Camgoz, N.C., Koller, O., Hadfield, S., Bowden, R.:
\newblock Sign language transformers: Joint end-to-end sign language
  recognition and translation.
\newblock In: CVPR. (2020)

\bibitem{Carreira2017}
Carreira, J., Zisserman, A.:
\newblock Quo vadis, action recognition? {A} new model and the {Kinetics}
  dataset.
\newblock In: CVPR. (2017)

\bibitem{Joze19msasl}
Joze, H.R.V., Koller, O.:
\newblock {MS-ASL}: {A} large-scale data set and benchmark for understanding
  american sign language.
\newblock In: BMVC. (2019)

\bibitem{Li19wlasl}
Li, D., Opazo, C.R., Yu, X., Li, H.:
\newblock Word-level deep sign language recognition from video: A new
  large-scale dataset and methods comparison.
\newblock In: WACV. (2019)

\bibitem{viitaniemi14}
Viitaniemi, V., Jantunen, T., Savolainen, L., Karppa, M., Laaksonen, J.:
\newblock S-pot – a benchmark in spotting signs within continuous signing.
\newblock In: LREC. (2014)

\bibitem{ong2014}
{Eng-Jon Ong}, {Koller}, O., {Pugeault}, N., {Bowden}, R.:
\newblock Sign spotting using hierarchical sequential patterns with temporal
  intervals.
\newblock In: CVPR. (2014)

\bibitem{Farhadi07}
Farhadi, A., Forsyth, D.A., White, R.:
\newblock Transfer learning in sign language.
\newblock In: CVPR. (2007)

\bibitem{Bilge19ZS}
Bilge, Y.C., Ikizler, N., Cinbis, R.:
\newblock Zero-shot sign language recognition: Can textual data uncover sign
  languages?
\newblock In: BMVC. (2019)

\bibitem{Motiian2017FewShotAD}
Motiian, S., Jones, Q., Iranmanesh, S.M., Doretto, G.:
\newblock Few-shot adversarial domain adaptation.
\newblock In: NeurIPS. (2017)

\bibitem{Zhang2019}
{Zhang}, J., {Chen}, Z., {Huang}, J., {Lin}, L., {Zhang}, D.:
\newblock Few-shot structured domain adaptation for virtual-to-real scene
  parsing.
\newblock In: ICCVW. (2019)

\bibitem{chang2019domain}
Chang, W.G., You, T., Seo, S., Kwak, S., Han, B.:
\newblock Domain-specific batch normalization for unsupervised domain
  adaptation.
\newblock In: CVPR. (2019)

\bibitem{li2020transferring}
Li, D., Yu, X., Xu, C., Petersson, L., Li, H.:
\newblock Transferring cross-domain knowledge for video sign language
  recognition.
\newblock In: CVPR. (2020)

\bibitem{Koller15cslr}
Koller, O., Forster, J., Ney, H.:
\newblock Continuous sign language recognition: Towards large vocabulary
  statistical recognition systems handling multiple signers.
\newblock Computer Vision and Image Understanding \textbf{141} (2015)  108--125

\bibitem{signum2008}
{von Agris}, U., {Knorr}, M., {Kraiss}, K.:
\newblock The significance of facial features for automatic sign language
  recognition.
\newblock In: 2008 8th IEEE International Conference on Automatic Face Gesture
  Recognition. (2008)

\bibitem{asllvid2008}
{Athitsos}, V., {Neidle}, C., {Sclaroff}, S., {Nash}, J., {Stefan}, A., {Quan
  Yuan}, {Thangali}, A.:
\newblock The american sign language lexicon video dataset.
\newblock In: CVPRW. (2008)

\bibitem{purdue06}
Wilbur, R.B., Kak, A.C.:
\newblock {Purdue RVL-SLLL} {A}merican sign language database.
\newblock School of Electrical and Computer Engineering Technical Report,
  TR-06-12, Purdue University, W. Lafayette, IN 47906. (2006)

\bibitem{chai2014devisign}
Chai, X., Wang, H., Chen, X.:
\newblock The devisign large vocabulary of chinese sign language database and
  baseline evaluations.
\newblock Technical report VIPL-TR-14-SLR-001. Key Lab of Intelligent
  Information Processing of Chinese Academy of Sciences (CAS), Institute of
  Computing Technology, CAS (2014)

\bibitem{schembri2013building}
Schembri, A., Fenlon, J., Rentelis, R., Reynolds, S., Cormier, K.:
\newblock Building the {B}ritish sign language corpus.
\newblock Language Documentation \& Conservation \textbf{7} (2013)  136--154

\bibitem{Cooper2009}
{Cooper}, H., {Bowden}, R.:
\newblock Learning signs from subtitles: A weakly supervised approach to sign
  language recognition.
\newblock In: CVPR. (2009)

\bibitem{Chung16b}
Chung, J.S., Zisserman, A.:
\newblock Signs in time: Encoding human motion as a temporal image.
\newblock In: Workshop on Brave New Ideas for Motion Representations, ECCV.
  (2016)

\bibitem{dietterich1997solving}
Dietterich, T.G., Lathrop, R.H., Lozano-P{\'e}rez, T.:
\newblock Solving the multiple instance problem with axis-parallel rectangles.
\newblock Artificial intelligence \textbf{89} (1997)  31--71

\bibitem{pfister2013large}
Pfister, T., Charles, J., Zisserman, A.:
\newblock Large-scale learning of sign language by watching tv (using
  co-occurrences).
\newblock In: BMVC. (2013)

\bibitem{Feng_2018_ECCV}
Feng, Y., Ma, L., Liu, W., Zhang, T., Luo, J.:
\newblock Video re-localization.
\newblock In: ECCV. (2018)

\bibitem{Yang_2018_CVPR}
Yang, H., He, X., Porikli, F.:
\newblock One-shot action localization by learning sequence matching network.
\newblock In: CVPR. (2018)

\bibitem{Cao_2020_CVPR}
Cao, K., Ji, J., Cao, Z., Chang, C.Y., Niebles, J.C.:
\newblock Few-shot video classification via temporal alignment.
\newblock In: CVPR. (2020)

\bibitem{oord2018representation}
Oord, A.v.d., Li, Y., Vinyals, O.:
\newblock Representation learning with contrastive predictive coding.
\newblock arXiv preprint arXiv:1807.03748 (2018)

\bibitem{wu2018unsupervised}
Wu, Z., Xiong, Y., Yu, S.X., Lin, D.:
\newblock Unsupervised feature learning via non-parametric instance
  discrimination.
\newblock In: CVPR. (2018)

\bibitem{miech2020end}
Miech, A., Alayrac, J.B., Smaira, L., Laptev, I., Sivic, J., Zisserman, A.:
\newblock End-to-end learning of visual representations from uncurated
  instructional videos.
\newblock In: CVPR. (2020)

\bibitem{signbslcom}
\url{https://www.signbsl.com/}:
\newblock (British sign language dictionary)

\bibitem{hu2019squeeze}
Hu, J., Shen, L., Albanie, S., Sun, G., Wu, E.:
\newblock Squeeze-and-excitation networks.
\newblock IEEE Transactions on Pattern Analysis and Machine Intelligence (2019)

\bibitem{Cao18}
Cao, Q., Shen, L., Xie, W., Parkhi, O.M., Zisserman, A.:
\newblock {VGGFace2}: A dataset for recognising faces across pose and age.
\newblock In: Proc. Int. Conf. Autom. Face and Gesture Recog. (2018)

\bibitem{cao2018openpose}
Cao, Z., Hidalgo, G., Simon, T., Wei, S.E., Sheikh, Y.:
\newblock Open{P}ose: realtime multi-person 2{D} pose estimation using {P}art
  {A}ffinity {F}ields.
\newblock In: arXiv preprint arXiv:1812.08008. (2018)

\bibitem{he2020momentum}
He, K., Fan, H., Wu, Y., Xie, S., Girshick, R.:
\newblock Momentum contrast for unsupervised visual representation learning.
\newblock In: CVPR. (2020)

\bibitem{SignumMcKee2000}
SignumMcKee, D., Kennedy, G.:
\newblock Lexical comparison of signs from {A}merican, {A}ustralian, {B}ritish
  and {N}ew {Z}ealand sign languages.
\newblock The signs of language revisited: {A}n anthology to honor {U}rsula
  {B}ellugi and {E}dward {K}lima (2000)

\bibitem{Aldersson2007}
Aldersson, R., McEntee-Atalianis, L.:
\newblock A lexical comparison of {I}celandic sign language and {D}anish sign
  language.
\newblock Birkbeck Studies in Applied Linguistics \textbf{2} (2007)

\bibitem{XieS3D}
Xie, S., Sun, C., Huang, J., Tu, Z., Murphy, K.:
\newblock Rethinking spatiotemporal feature learning for video understanding.
\newblock In: ECCV. (2018)

\bibitem{NEURIPS2019_9015}
Paszke, A., Gross, S., Massa, F., Lerer, A., Bradbury, J., Chanan, G., Killeen,
  T., Lin, Z., Gimelshein, N., Antiga, L., Desmaison, A., Kopf, A., Yang, E.,
  DeVito, Z., Raison, M., Tejani, A., Chilamkurthy, S., Steiner, B., Fang, L.,
  Bai, J., Chintala, S.:
\newblock Pytorch: An imperative style, high-performance deep learning library.
\newblock In Wallach, H., Larochelle, H., Beygelzimer, A., d'Alch\'{e} Buc, F.,
  Fox, E., Garnett, R., eds.: Advances in Neural Information Processing Systems
  32.
\newblock Curran Associates, Inc. (2019)  8024--8035

\end{thebibliography}

% Appendix
\newpage
\noindent \textbf{\large APPENDIX}
\bigskip
\renewcommand{\thefigure}{A.\arabic{figure}}
% \thesection instead of A would make it A.1, B.1...
\setcounter{figure}{0} 
\renewcommand{\thetable}{A.\arabic{table}}
\setcounter{table}{0} 

\appendix

This appendix provides additional qualitative (Sec.~\ref{app:sec:qualitative}) and experimental results (Sec.~\ref{app:sec:additionalexp}),
as well as detailed explanations of the training of our Watch-Read-Lookup framework (Sec.~\ref{app:sec:details}).

\section{Qualitative Results}\label{app:sec:qualitative}
Please watch our % supplemental video {\texttt{qualitative.mp4}}
video in the project webpage\footnote{\url{https://www.robots.ox.ac.uk/~vgg/research/bsldict/}}
to see qualitative results of our model in action. We  illustrate the sign spotting task, as well as the specific applications considered in the main paper: sign variant identification, densification of annotations, and ``faux amis'' identification between languages.

\section{Additional Experiments}\label{app:sec:additionalexp}
In this section, we present complementary experimental results
to the main paper.
We report the variance of the results
over multiple random seeds (Sec.~\ref{app:subsec:variance}),
the effect of class-balancing (Sec.~\ref{app:subsec:balancing}),
domain-specific layers (Sec.~\ref{app:subsec:domainspecific}),
language-aware negative sampling (Sec.~\ref{app:subsec:languageaware}),
sliding window stride at test time (Sec.~\ref{app:subsec:stride}), 
the mouthing score threshold (Sec.~\ref{app:subsec:mouthingthres}),
and the trunk network architecture
(Sec.~\ref{app:subsec:s3d}).

% ======================= Variance of results ======================= 
\subsection{Variance of results}\label{app:subsec:variance}
We repeat the experiments in
\if\sepappendix1{Tables~2~and~3 of the main paper,}
\else{Tables~\ref{tab:loss}~and~\ref{tab:vocab} of the main paper,}
\fi
with multiple random seeds for each model and report means and standard deviations
in Tab.~\ref{app:tab:loss} and Tab.~\ref{app:tab:vocab} to provide a measure of the
variance of the results.
We observe that the results are consistent with those reported in the main paper.

\begin{table}
    \setlength{\tabcolsep}{8pt}
    \centering
    \resizebox{0.99\linewidth}{!}{
        \begin{tabular}{lcc|cc}
            \toprule
            & \multicolumn{2}{c|}{Train (1064)} & \multicolumn{2}{c}{Train (800)} \\
            \midrule
            & \multicolumn{2}{c|}{Seen (264)} & \multicolumn{2}{c}{Unseen (264)} \\
            Supervision & mAP & R@5 & mAP & R@5 \\
            \midrule
            % Classification & 36.75 & 40.15 & 10.28 & 14.19 \\
            % Classification & 37.44 & 40.00 & 9.83 & 11.76 \\
            % Classification & 37.19 & 38.88 & 10.89 & 14.05 \\
            % \midrule
            Classification & 37.13 $\pm$ 0.29 & 39.68 $\pm$ 0.57 & 10.33 $\pm$ 0.43 & 13.33 $\pm$ 1.11 \\
            % \midrule
            % NCE & 42.52 & 53.54 & 10.88 & 14.23 \\
            % NCE & 44.23 & 52.50 & 11.90 & 15.20 \\
            % NCE & 44.03 & 51.72 & 11.41 & 14.86 \\
            % \midrule
            NCE & 43.59 $\pm$ 0.76 & 52.59 $\pm$ 0.75 & 11.40 $\pm$ 0.42 & 14.76 $\pm$ 0.40 \\
            % \midrule
            % Watch-Lookup & 43.65 & 53.03 & 11.05 & 14.62 \\
            % Watch-Lookup & 44.78 & 58.32 & 11.33 & 15.65 \\
            % Watch-Lookup & 45.72 & 55.17 & 10.67 & 14.82 \\
            % \midrule
            Watch-Lookup & 44.72 $\pm$ 0.85 & 55.51 $\pm$ 2.17 & 11.02 $\pm$ 0.27 & 15.03 $\pm$ 0.45 \\
            % \midrule
            % Watch-Read-Lookup & 48.11 & 58.71 & 13.69 & 17.79 \\
            % Watch-Read-Lookup & 47.65 & 61.89 & 14.24 & 19.31 \\
            % Watch-Read-Lookup & 48.04 & 61.67 & 16.65 & 22.46 \\
            % \midrule
            Watch-Read-Lookup & \textbf{47.93} $\pm$ 0.20 & \textbf{60.76} $\pm$ 1.45 & \textbf{14.86} $\pm$ 1.29 & \textbf{19.85} $\pm$ 1.94 \\
            \bottomrule
        \end{tabular}
    }
    \caption{\textbf{Variance of the results with multiple random seeds:} We repeat the
        \if\sepappendix1{Table~2 experiments of the main paper,}
        \else{Table~\ref{tab:loss} experiments of the main paper,}
        \fi
        with three random seeds for each model and report the mean and the standard deviation.
    }
    \label{app:tab:loss}
\end{table}
\begin{table}[h!]
    \setlength{\tabcolsep}{8pt}
    \centering
    \resizebox{0.9\linewidth}{!}{
        \begin{tabular}{llcc}
            \toprule
            Supervision & Dictionary Vocab & mAP & R@5 \\
            \midrule
            % Watch-Read-Lookup & 800 training vocab & 13.69 & 17.79 \\ 
            % Watch-Read-Lookup & 800 training vocab & 14.24 & 19.31 \\ 
            % Watch-Read-Lookup & 800 training vocab & 16.65 & 22.46 \\
            % \midrule
            Watch-Read-Lookup & 800 training vocab & 14.86 $\pm$ 1.29 & 19.85 $\pm$ 1.94 \\
            % \midrule
            % Watch-Read-Lookup & 9k full vocab & 15.39 & 20.87 \\
            % Watch-Read-Lookup & 9k full vocab & 15.58 & 22.61 \\
            % Watch-Read-Lookup & 9k full vocab & 16.48 & 21.53 \\
            % \midrule
            Watch-Read-Lookup & 9k full vocab & \textbf{15.82} $\pm$ 0.48 & \textbf{21.67} $\pm$ 0.72 \\
            \bottomrule
        \end{tabular}
    }
    \caption{\textbf{Variance of the results with multiple random seeds:} We repeat the
        \if\sepappendix1{Table~3 experiments of the main paper}
        \else{Table~\ref{tab:vocab} experiments of the main paper}
        \fi
        with three random seeds for each model and report the mean and the standard deviation.
    }
    \label{app:tab:vocab}
\end{table}
\begin{table}[h!]
    \setlength{\tabcolsep}{8pt}
    \centering
    \resizebox{0.6\linewidth}{!}{
        \begin{tabular}{llcc}
            \toprule
            Class-balancing & Batch size & mAP & R@5 \\
            \midrule
            \xmark & 512 & 41.65 & 54.73 \\
            \xmark & 1024 & 42.07 & 54.25 \\
            \xmark & 2048 & 43.14 & 54.28 \\
            \midrule
            \cmark & 512 & 43.65 & 53.03 \\
            \cmark & 1024 & 43.55 & 54.20 \\
            \bottomrule
        \end{tabular}
    }
    \caption{\textbf{Class-balancing:}
    In the main paper, we class-balance our minibatches by including
    one sample per word from the labelled continuous sequences, thus maximizing the number of negatives
    within a batch.
    Here, we investigate removing such class-balancing constraint.
    In that case, we make sure we do not mark samples
    with the same labels as negatives, instead we discard them.
    We experiment with various batch sizes, also going beyond
    the total number of classes (2048). We observe that
    the performance is not significantly affected
    by these changes.
    (training on the full 1064 vocabulary with Watch-Lookup)
    }
    \label{app:tab:balancing}
\end{table}

% ======================= Class-balanced sampling ======================= 
\subsection{Class-balanced sampling}\label{app:subsec:balancing}
As described in the main paper, we construct each batch by maximizing the number
of negative pairs. To this end, we include one labelled sample per word when sampling
continuous sequences, i.e., class-balancing the minibatches. Thus,
all but one of the labelled samples in the batch can be used as negatives
for a given dictionary bag corresponding to a labelled sample.
Note that this approach limits the batch size to be less than or equal to the
number of sign classes.
Tab.~\ref{app:tab:balancing} experiments with the sampling strategy.
We observe that the performance is not significantly
different with/without class-balanced sampling
for various batch sizes.

% ======================= Domain-specific layers ======================= 
\subsection{Domain-specific layers}\label{app:subsec:domainspecific}
As noted in the main paper, the videos from the continuous signing and from the dictionaries
differ significantly, e.g., continuous signing data is faster than the dictionary signing, and is co-articulated whereas the dictionary has isolated signs.
Given such a domain gap, we explore whether it is beneficial to learn
domain-specific MLP layers: one for the continuous, and one for the dictionary.
Tab.~\ref{app:tab:domainspecific} presents a comparison between
domain-specific layers versus shared parameters. We do not observe any
gains from such separation. Therefore, we keep a single MLP for both domains
for simplicity.

\begin{table}[t]
    \setlength{\tabcolsep}{8pt}
    \centering
    \resizebox{0.6\linewidth}{!}{
        \begin{tabular}{lcc}
            \toprule
            Domain-specific layers  & mAP & R@5 \\
            \midrule
            \cmark & 43.58 & 53.54 \\
            \xmark & 43.65 & 53.03 \\
            \bottomrule
        \end{tabular}
    }
    \caption{\textbf{Domain-specific layers:} We experiment with separating the MLP layers
    to be specific to the continuous and isolated domains. We do not observe
    any significant difference in performance and therefore adopt a shared MLP
    for simplicity in all experiments. (Training on the full 1064 vocabulary with Watch-Lookup)
    }
    \label{app:tab:domainspecific}
\end{table}

% ======================= Language-aware negative sampling ======================= 
\subsection{Language-aware negative sampling}\label{app:subsec:languageaware}
Working with a large vocabulary of words brings the additional
challenge of handling synonyms. We consider two types of similarities.
First, two different categories in the \bsldict{} sign dictionary may belong
to the same sign category if the corresponding English words are synonyms.
Second, the meta-data we have collected with the \bsldict{} dataset
provides similarity labels between sign categories, which may be used
to group certain signs.
In this work, we have largely ignored this issue by associating
each sign to a single word. This results in constructing
negative pairs for two identical signs such as `happy' and `content'.
Here, we explore whether it is beneficial to discard
such pairs during training, instead of marking them as negatives.
Tab.~\ref{app:tab:languageaware}
reports the results. We observe marginal gains with discarding synonyms.
However, given the insignificant difference, we
do not make such separation in other experiments for simplicity.

\begin{table}[t]
    \setlength{\tabcolsep}{8pt}
    \centering
    \resizebox{0.7\linewidth}{!}{
        \begin{tabular}{lcc}
            \toprule
            Negative sampling  & mAP & R@5 \\
            \midrule
            Discarding English synonyms & 43.27 & 54.24 \\
            Discarding Sign synonyms & 45.03 & 54.19 \\
            Keeping all & 43.65 & 53.03 \\
            \bottomrule
        \end{tabular}
    }
    \caption{\textbf{Language-aware negative sampling:} We explore the use of external knowledge
    such as English synonyms or the meta-data of the dictionary denoting similar sign categories.
    We experiment with discarding such similar word pairs, excluding them from both
    positive and negative pairs. The last row instead marks any pair
    as negative if their corresponding words are not identical. We observe only marginal
    gains with the use of external knowledge about the languages. (Training on the full 1064 vocabulary with Watch-Lookup)
    }
    \label{app:tab:languageaware}
\end{table}

\subsection{Effect of the sliding window stride}\label{app:subsec:stride}
As explained in the main paper,
at test time, we extract features from the continuous signing
sequence using a sliding window approach with 1 frame as the stride parameter.
Our window size is 16 frames, i.e., the number of input frames for the I3D
feature extractor. Here, we investigate the effect of the stride parameter.
We apply a stride of 8 frames as a comparison. Tab.~\ref{app:tab:stride}
shows that a stride of 1 frame is critical to perform precise
sign spotting. This can be explained by the fact that sign duration is typically
between 7-13 frames (but can be shorter)~\cite{pfister2013large} in continuous signing video, and a stride of 8 may skip
the most discriminative moment.

\begin{table}[t]
    \setlength{\tabcolsep}{8pt}
    \centering
    \resizebox{0.4\linewidth}{!}{
        \begin{tabular}{lcc}
            \toprule
            Stride & mAP & R@5 \\
            \midrule
            8 & 38.46 & 47.38 \\
            1 & \textbf{43.65} & \textbf{53.03} \\
            \bottomrule
        \end{tabular}
    }
    \caption{\textbf{Stride parameter of sliding window:} A small stride
    at test time, when extracting embeddings from the continuous signing video,
    allows us to temporally localise the signs more precisely. The window size is
    16 frames and the typical co-articulated sign duration is 7-13 frames (at 25 fps).
    (testing 1064-class model trained with Watch-Lookup)
    }
    \label{app:tab:stride}
\end{table}

\subsection{Mouthing confidence threshold at training}\label{app:subsec:mouthingthres}
The sparse annotations from the \bslonek{} dataset are obtained
by running a visual keyword spotting method based on mouthing cues.
Therefore, the dataset provides a confidence value associated with
each label ranging between $0.5$ and $1.0$. Similar to \cite{bsl1k2020},
we experiment with different thresholds to determine the training set.
Lower thresholds result in a noisier but larger training set.
From Tab.~\ref{app:tab:mouthingthres}, we conclude that
$0.5$ mouthing confidence threshold performs the best.
This is in accordance with the conclusion from \cite{bsl1k2020}.

\begin{table}
    \setlength{\tabcolsep}{8pt}
    \centering
    \resizebox{0.8\linewidth}{!}{
        \begin{tabular}{llcc}
            \toprule
            Mouthing confidence & Training size & mAP & R@5 \\
            \midrule
            0.9 & 10K & 37.55 & 47.54 \\ % 10555
            0.8 & 21K & 39.49 & 48.84 \\ % 20929
            0.7 & 33K & 41.87 & 51.15 \\ % 32971
            0.6 & 49K & 42.44 & 52.42 \\ % 49303
            0.5 & 78K & \textbf{43.65} & \textbf{53.03} \\ % 78211
            \bottomrule
        \end{tabular}
    }
    \caption{\textbf{Mouthing confidence threshold:} The results suggest that lower confidence automatic annotations of \bslonek{} provide better training, by increasing the amount of data (training on the full 1064 vocabulary with Watch-Lookup).
    }
    \label{app:tab:mouthingthres}
\end{table}

\subsection{Trunk network architecture: S3D vs I3D} \label{app:subsec:s3d}
As shown in Tab~\ref{app:tab:s3d},
we compare two popular architectures
for computing video representations. We have used
I3D~\cite{Carreira2017} in all our experiments.
Here, we also train a 1064-way classification
with the S3D architecture~\cite{XieS3D}
on \bslonek{} as in \cite{bsl1k2020} for sign
language recognition. We do not observe
improvements with S3D (in practice we found that it overfit the training set to a greater degree); therefore, we use
an I3D trunk. Note that the hyperparameters
(e.g., learning rate)
are tuned
for I3D and kept the same for S3D.

\begin{table}
    \setlength{\tabcolsep}{8pt}
    \centering
    \resizebox{0.7\linewidth}{!}{
        \begin{tabular}{l|cccc}
            \toprule
            & \multicolumn{2}{c}{per-instance} & \multicolumn{2}{c}{per-class} \\
            Training data & top-1 & top-5 & top-1 & top-5  \\
            \midrule
            S3D & 64.76	& 81.88	& 46.27 & 63.71 \\
            I3D~\cite{bsl1k2020} & \textbf{75.51} & \textbf{88.83} & \textbf{52.76} & \textbf{72.14} \\
            \bottomrule
        \end{tabular}
    }
    \caption{
    \textbf{Trunk network architecture:} We compare
    I3D~\cite{Carreira2017} with the S3D~\cite{XieS3D} architecture
    for the task of sign language recognition,
    in a comparable setup to~\cite{bsl1k2020}.
    We use the last 20 frames before the mouthing
    annotations with confidence above 0.5. We do not
    obtain gains with the S3D architecture; therefore,
    we use I3D in all the experiments to compute video
    features.
    }
    \label{app:tab:s3d}
\end{table}

\section{Training Details}\label{app:sec:details}
In this section, we cover architectural details (Sec.~\ref{app:subsec:arch}),
a detailed formulation of our positive/negative
bag sampling strategy (Sec.~\ref{app:subsec:math}) and a brief description of the infrastructure used to perform the experiments in the main paper (Sec.~\ref{app:subsec:infra}).

\begin{figure}[t]
    \centering
    \includegraphics[width=.75\textwidth]{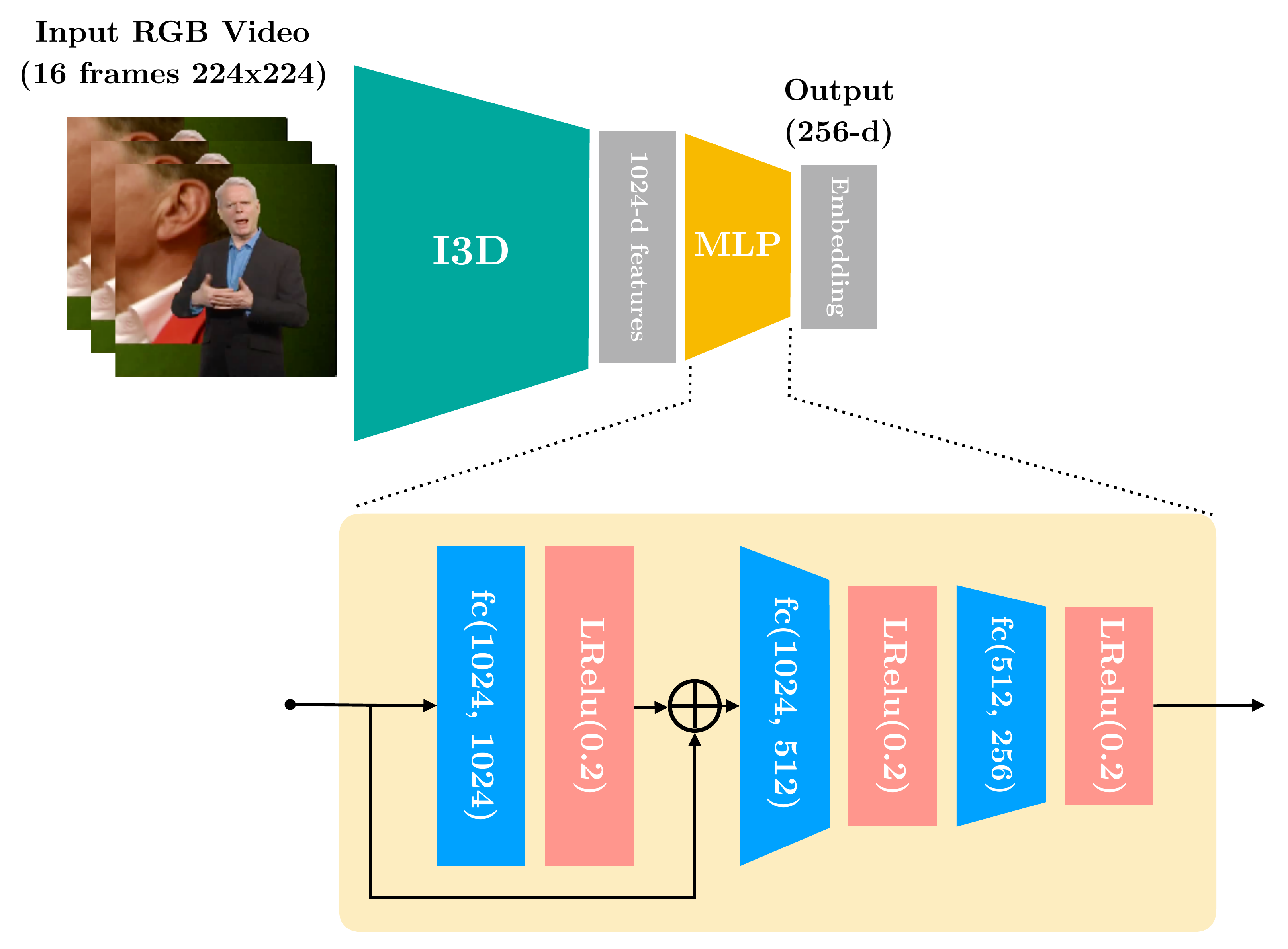}
    \caption{\textbf{MLP architecture:} We detail the layers of our embedding architecture.
    We freeze the I3D trunk and use it as a feature extractor.
    We only train the MLP layers with our loss formulation in the proposed framework. The same layers (and parameters) are used both for the dictionary video inputs and the continuous signing video inputs.}
    \label{app:fig:arch}
\end{figure}

\subsection{Architectural details}\label{app:subsec:arch}
As explained in the main paper, our sign embeddings correspond
to the output of a two-stage architecture: (i) an I3D trunk, and (ii) a three-layer MLP.
We first train the I3D on both labelled continuous video clips
and the dictionary videos jointly. We then freeze the I3D trunk
and use it as a feature extractor. We only train the MLP layers with our loss formulation in the Watch-Read-Lookup framework. \\

\noindent\textbf{I3D trunk.}
We first train the I3D parameters only with the \bslonek{} annotated clips
that have mouthing confidences more than 0.5. For 1064-class training,
we use the model from \cite{bsl1k2020} provided by the authors; for 800-class training, we perform
our own training, also first pretraining with pose distillation.

We then \textit{re-initialise the batch normalization layers}
(as noted in
\if\sepappendix1{Sec.~2 of the main paper).}
\else{Sec.~\ref{sec:related} of the main paper).}
\fi
We fine-tune the model jointly on \bslonek{} annotated clips (the ones
with mouthing confidence more than 0.8)
and \bsldict{} samples. The sampling frequency for the two data sources are balanced.
In the I3D classification pretraining phase, we treat each dictionary video
independently with its corresponding label.
We observe that the 1064-way classification performance
on the \textit{training} dictionary videos remain at 48.09\% per-instance top-1
accuracy without the batch normalization re-initialization, as opposed
to 78.94\%. We also experimented with domain-specific batch normalization layers~\cite{chang2019domain},
but the training accuracy for the dictionary videos was still low (62.73\%).

As detailed in
\if\sepappendix1{Sec.~3.2 of the main paper,}
\else{Sec.~\ref{subsection:implementation} of the main paper,}
\fi
we subsample the dictionary
videos to roughly match their speed to the continuous signing videos.
This subsampling includes \textit{a random shift and a random fps}. We observe
a decrease of 6.68\% in the training dictionary classification accuracy
if we instead sample 16 consecutive frames from the original
temporal resolution, which is not sufficient to capture the full extent of
a sign because one dictionary video is 56 frames on average. \\

\noindent\textbf{MLP.} Fig.~\ref{app:fig:arch} illustrates
the layers considered for our MLP architecture. It consists of 3 fully connected
layers with LeakyRelu activations between them. The first linear layer also has a
residual connection on the 1024-dimensional input features. We then reduce
the dimensionality gradually to 512 and 256 for efficient training and testing. \\

\subsection{Positive/Negative bag sampling formulations}\label{app:subsec:math}

In the main paper, we described two approaches for sampling positive/negative MIL bags in
\if\sepappendix1{Sec.~3.1.}
\else{Sec.~\ref{subsection:mil}.}
\fi
Due to space constraints, the sampling mechanisms were described at a high-level.  Here, we provide more precise definitions of each bag.  In addition to the set notation below, we include in the supplementary material, the loss implementation as a PyTorch~\cite{NEURIPS2019_9015} function in \texttt{code/loss.py}, together with a sample
input ({\texttt{code/sample\_inputs.pkl}}) comprising embedding outputs from the MLP
for continuous and dictionary videos. % \footnote{Full source-code and pretrained models will be made available with the final paper.}.

As noted in the main paper, we do not have access to positive pairs because: (1) for the segments of videos in $\mathcal{S}$ that are annotated (i.e. $(x_k, v_k) \in \mathcal{M}$), we have a set of potential sign variations represented in the dictionary (annotated with the common label $v_k$), rather than a single unique sign; (2) since $\mathcal{S}$ provides only weak supervision, even when a word is mentioned in the subtitles we do not know where it appears in the continuous signing sequence (if it appears at all). 
These ambiguities motivate a Multiple Instance Learning~\cite{dietterich1997solving} (MIL) objective. Rather than forming positive and negative pairs, we instead form positive \textit{bags} of pairs, $\mathcal{P}^{\text{bags}}$, in which we expect at least one segment from a video from $\mathcal{S}$ (or a video from $\mathcal{M}$ when labels are available) and a video $\mathcal{D}$ to contain the same sign, and negative bags of pairs, $\mathcal{N}^{\text{bags}}$, in which we expect no pair of video segments from $\mathcal{S}$ (or $\mathcal{M}$) and $\mathcal{D}$ to contain the same sign.  To incorporate the available sources of supervision into this formulation, we consider two categories of positive and negative bag formations, described next. Each bag is formulated as a set of paired indices---the first value indexes into the collections of continuous signing videos (either $\mathcal{S}$ or $\mathcal{M}$, depending on context) and the second value indexes into the set of dictionary videos contained in $\mathcal{D}$. \\

\noindent \textbf{Watch and Lookup: using sparse annotations and dictionaries}. In the first formulation, \textit{Watch-Lookup}, we only make use of $\mathcal{D}$ and $\mathcal{M}$ (and not $\mathcal{S}$) to learn the data representation $f$.  We define positive bags in two ways: (1) by anchoring on the labelled segment  

\begin{align}
    \mathcal{P}_{\text{watch,lookup}}^{\text{bags(seg)}} = \{\{i\} \times B_i: (x_i^\mathcal{M}, v_i^\mathcal{M}) \in \mathcal{M}, (x_j^\mathcal{D}, v_j^\mathcal{D}) \in \mathcal{D}, B_i = \{j : v_j^\mathcal{D} =  v_i^\mathcal{M}\}\}
\end{align}

\begin{figure}[t]
    \centering
    \includegraphics[width=\textwidth]{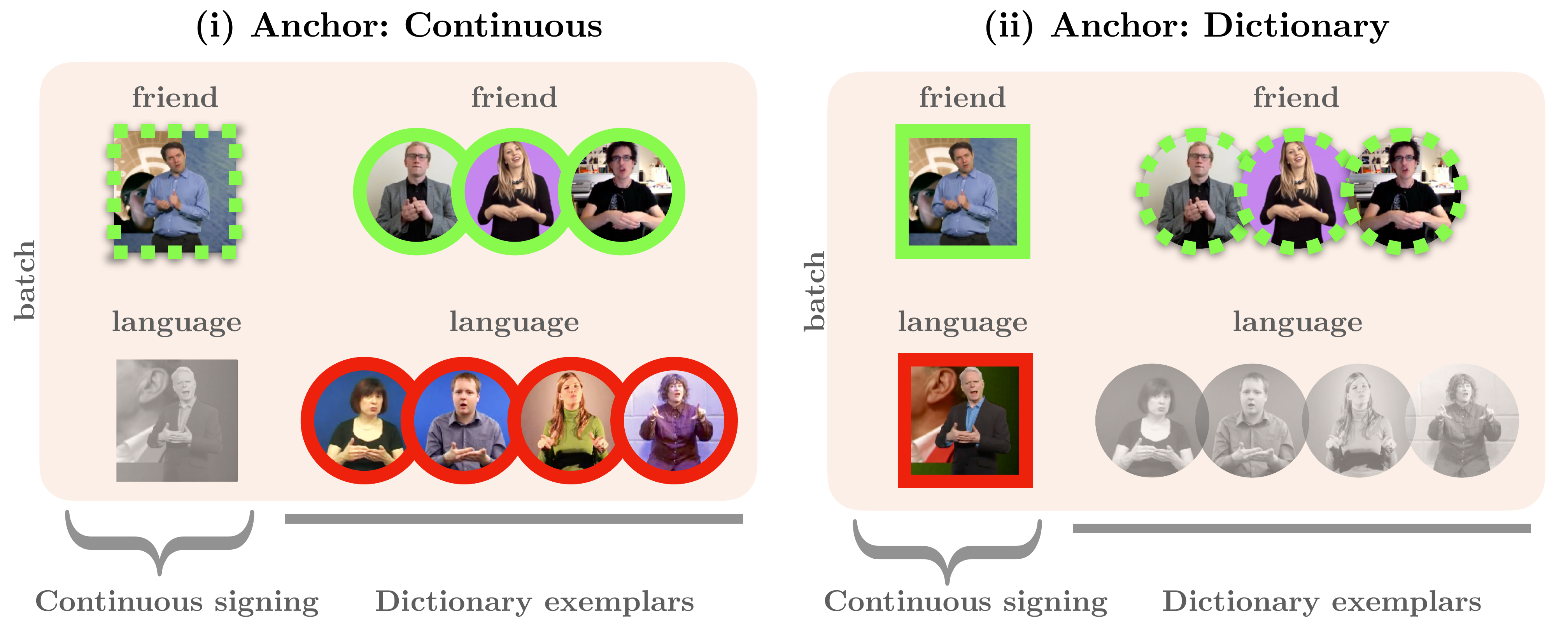}
    \caption{\textbf{Watch-Lookup:} We illustrate the batch formation and positive/negative sampling
    for the simplified version of our framework which
    is not using the subtitles, but only performing
    Watch-Lookup.
    We define two sets of positive/negative pairs, anchoring at a different position in each case.
    Anchor is denoted with dashed lines, positive samples
    with solid green, negative samples with solid red lines. Gray samples are discarded.
    (i) anchors at a labelled continuous video, making the dictionary samples for the labelled word a positive bag, and all other dictionary samples in the batch a negative bag.
    (ii) anchors at a bag of dictionary samples, making the corresponding continuous labelled video positive,
    and all others in the batch negatives.
    We refer to Fig~\ref{app:fig:sampling_mil2nce} for the illustration
    of our Watch-Read-Lookup extension.
    }
    \label{app:fig:sampling_milnce}
\end{figure}

\noindent i.e. each bag consists of a labelled temporal segment and the set of sign variations of the corresponding word in the dictionary (illustrated in Fig.~\ref{app:fig:sampling_milnce} (i), top row), or by (2) anchoring on the dictionary samples that correspond to the labelled segment, to define a second set $\mathcal{P}_{\text{watch,lookup}}^{\text{bags(dict)}}$, which takes a mathematically identical form to $\mathcal{P}_{\text{watch,lookup}}^{\text{bags(seg)}}$ (i.e. each bag consists of the set of sign variations of the word in the dictionary that corresponds to a given labelled temporal segment, illustrated in Fig.~\ref{app:fig:sampling_milnce} (ii), top row). The key assumption in both cases is that each labelled segment matches \textit{at least one} sign variation in the dictionary. Negative bags can be constructed by (1) anchoring on labelled segments and selecting dictionary examples corresponding to different words (Fig.~\ref{app:fig:sampling_milnce} (i), red examples); (2) anchoring on the dictionary set for a given word and selecting labelled segments of a different word  (Fig.~\ref{app:fig:sampling_milnce} (ii), red example).
These sets manifest as

\begin{align}
    \mathcal{N}_{\text{watch,lookup}}^{\text{bags(seg)}} = \{\{i\} \times B_i: (x_i^\mathcal{M}, v_i^\mathcal{M}) \in \mathcal{M}, (x_j^\mathcal{D}, v_j^\mathcal{D}) \in \mathcal{D}, B_i = \{j : v_j^\mathcal{D} \neq v_i^\mathcal{M} \}\}
\end{align}

\noindent for the former and as

\begin{align}
    \mathcal{N}_{\text{watch,lookup}}^{\text{bags(dict)}} = \{A_i \times B_i: &A_i = \{l: x_l, x_i \subseteq x_k, (x_k, s_k) \in \mathcal{S}, x_l \cap x_i = \emptyset\} \\ \nonumber & B_i = \{j : v_j^\mathcal{D} \neq v_i^\mathcal{M} \}, (x_i^\mathcal{M}, v_i^\mathcal{M}) \in \mathcal{M}, (x_j^\mathcal{D}, v_j^\mathcal{D}) \in \mathcal{D}
    \}.
\end{align}

\noindent for the latter. The complete set of positive and negative bags is formed via the unions of these collections:

\begin{align}
\mathcal{P}_{\text{watch,lookup}}^{\text{bags}} \triangleq \mathcal{P}_{\text{watch,lookup}}^{\text{bags(seg)}} \cup \mathcal{P}_{\text{watch,lookup}}^{\text{bags(dict)}}
\end{align}

\noindent and

\begin{align}
    \mathcal{N}_{\text{watch,lookup}}^{\text{bags}} \triangleq \mathcal{N}_{\text{watch,lookup}}^{\text{bags(seg)}} \cup \mathcal{N}_{\text{watch,lookup}}^{\text{bags(dict)}}.
\end{align}

\begin{figure}
    \centering
\subfloat[
\textbf{Input:} We illustrate an example minibatch formation
for our Watch-Read-Lookup framework. We sample continuous
videos with only one labelled segment, which we refer to as the `foreground' word (e.g., \textit{friend}, \textit{language}). Each continuous video has a subtitle, which we use to sample additional words for which we do not have continuous signing labels, (`background' words), e.g. \textit{name} and \textit{what} for \textit{``what is your friend's name?''}. We sample all the dictionary videos corresponding to these words. Each word has multiple dictionary instances grouped into overlapping circles.
]
    {\makebox[12cm][c]{
    \includegraphics[width=0.9\textwidth]{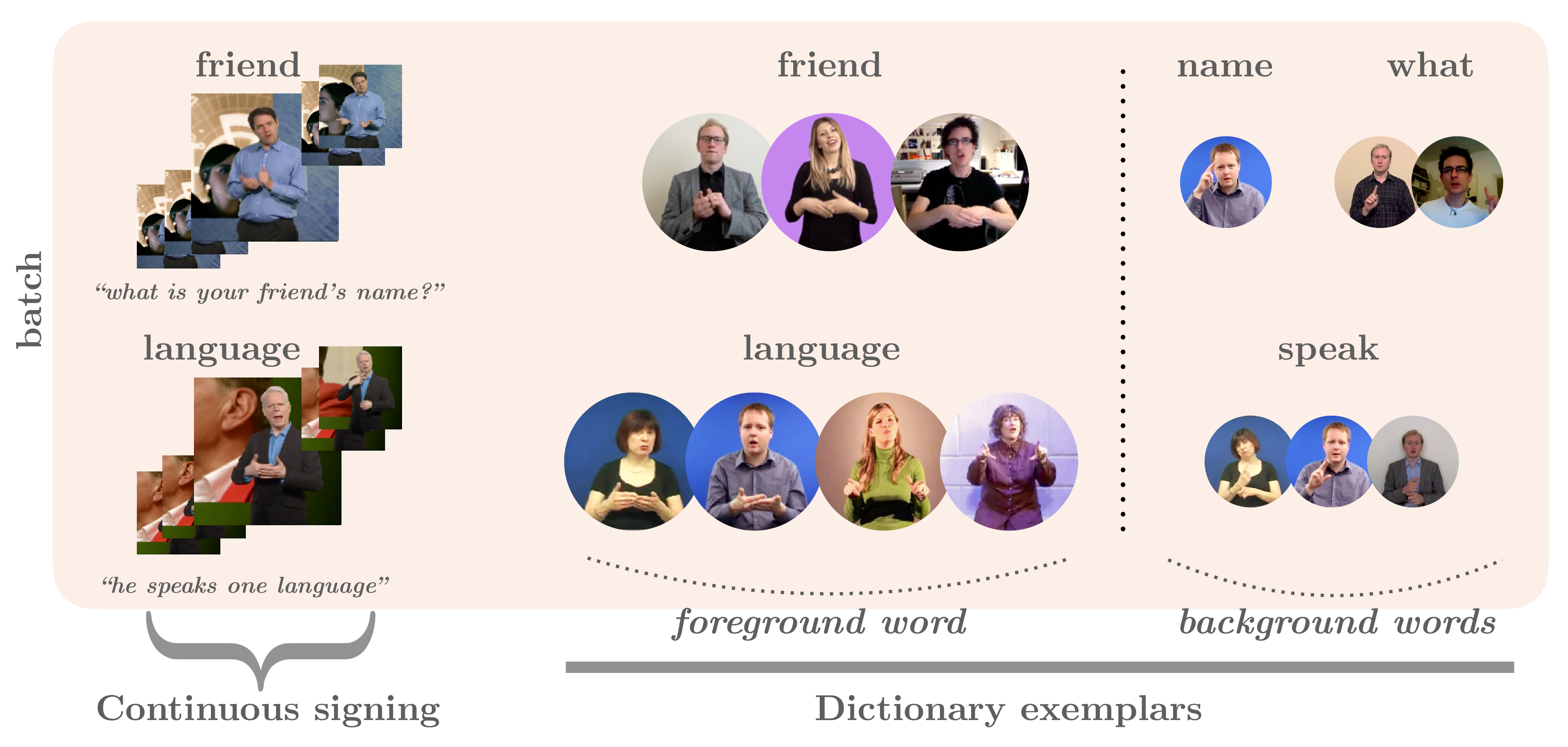}}} \\
\subfloat[
\textbf{Sampling positive/negative pairs:}
We anchor at 4 different positions within the batch to determine the pairs. Anchors are denoted with dashed lines, positive samples with solid green, negative samples with solid red lines. Gray samples are discarded.
For example, (iii) anchoring at the continuous background
marks the dictionary video for \textit{name} positive, because
it appears in the subtitle, but it is not within the annotated
temporal window. All other dictionary samples \textit{friend, language, speak} become negative to this anchor.
We repeat this for each dictionary background, i.e.,
marking \textit{what} as positive.
See text for detailed explanations on each case.
We also provide a %supplementary video animation {\texttt{sampling.mp4}}
video animation at our project page
to show all possible positive/negative pairs
for cases (i) to (iv).
]{
    \includegraphics[width=\textwidth]{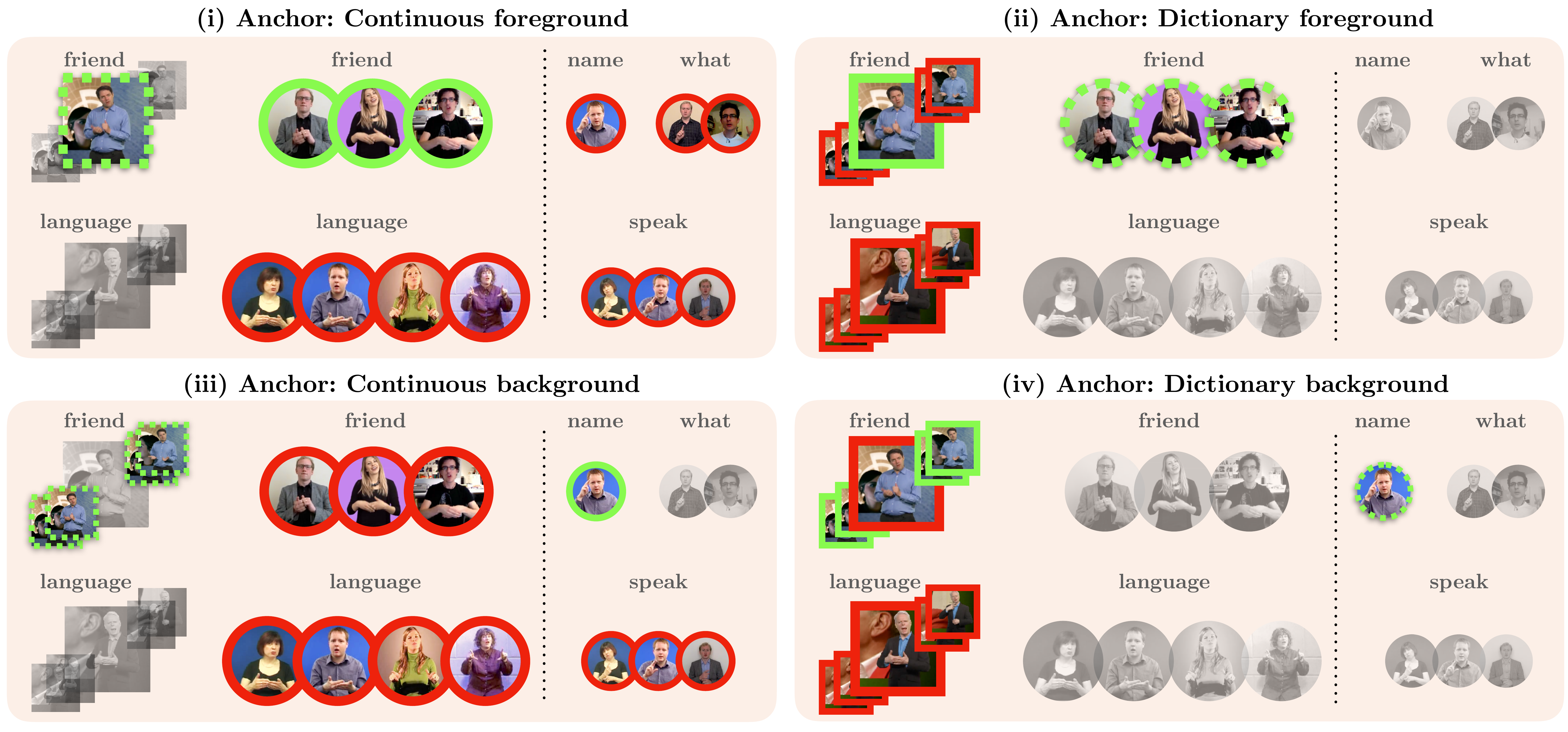}}
    \caption{\textbf{Watch-Read-Lookup in detail.}
    }
    \label{app:fig:sampling_mil2nce}
\end{figure}

\noindent \textbf{Watch, Read and Lookup}. The \textit{Watch-Lookup} bag formulation defined above has a significant limitation: the data representation, $f$, is not encouraged to represent signs beyond the initial vocabulary represented in $\mathcal{M}$.  We therefore look at the subtitles present in $\mathcal{S}$ (which contain words beyond $\mathcal{M}$) in addition to $\mathcal{M}$ to construct bags. To do so, we introduce an additional piece of terminology---when considering a subtitled video for which only one segment is labelled, we use the term \say{foreground} to refer to the subtitle word that corresponds to the label, and \say{background} for words which do not possess labelled segments in the video. Similarly to \textit{Watch-Lookup}, we can construct positive bags,  $\mathcal{P}_{\text{watch,lookup}}^{\text{bags}}$ (Fig.~\ref{app:fig:sampling_mil2nce} (i) and (ii), top rows) which correspond to the use of foreground subtitle words.  However, these can now by extended by (a) anchoring on a background segment in the continuous footage and find candidate matches in the dictionary among all possible matches for the subtitles words (Fig.~\ref{app:fig:sampling_mil2nce} (iii), top row) and (b) anchoring on dictionary entries for background subtitle words (Fig.~\ref{app:fig:sampling_mil2nce} (iv), top row). Formally, let Tokenize$(\cdot): \mathcal{S} \rightarrow \mathcal{V}_\mathfrak{L}$ denote the function which extracts words from the subtitle that are present in the vocabulary: Tokenize$(s) \triangleq \{ w \in s: w \in \mathcal{V}_\mathfrak{L}\}$. Then define background segment-anchored positive bags as: 
\nopagebreak
\begin{align}
\mathcal{P}_{\text{watch,read,lookup}}^{\text{bags(seg-back)}} = \{\{i\} \times B_i: \exists (x_k, s_k) \in \mathcal{S} \text{ s.t } x_i \subseteq x_k, (x_j^\mathcal{D}, v_j^\mathcal{D}) \in \mathcal{D}, \\ \nonumber B_i = \{j : v_j^\mathcal{D} \in \text{Tokenize}(s_k)\}, (x_i, v_i) \notin \mathcal{M}\}
\end{align}

\noindent i.e. each bag contains a background segment from the continuous signing which is paired with all dictionary segments whose labels match any token from the corresponding subtitle sentence (visualised as the top row of Fig.~\ref{app:fig:sampling_mil2nce} (iii)). Next, we define dictionary-anchored positive background bags as follows:
\begin{align}
    \mathcal{P}_{\text{watch,read,lookup}}^{\text{bags(dict-back)}} = \{A_i \times B_i: (x_i^\mathcal{D}, v_i^\mathcal{D}) \in \mathcal{D}, A_i = \{j : v_i^\mathcal{D} \in \text{Tokenize}(s_k), \\ \nonumber (x_k, s_k) \in \mathcal{S}, x_j \subseteq x_k, (x_j, v_j) \notin \mathcal{M}\}, B_i = \{l: v_l^\mathcal{D} = v_i^\mathcal{D}\}\}
\end{align}

\noindent i.e. the bags contain all pairwise combinations of dictionary entries for a given word and segments in continuous signing whose subtitle contains that background word (visualised as top row of Fig.~\ref{app:fig:sampling_mil2nce} (iv)). 
We combine these bags with the \textit{Watch-Lookup} positive bags to maximally exploit the available supervisory signal for positives:

\begin{align}
\mathcal{P}_{\text{watch,read,lookup}}^{\text{bags}} = \mathcal{P}_{\text{watch,lookup}}^{\text{bags}} \cup \mathcal{P}_{\text{watch,read,lookup}}^{\text{bags(seg-back)}} \cup \mathcal{P}_{\text{watch,read,lookup}}^{\text{bags(dict-back)}}.
\end{align}

\noindent To counterbalance the positives, we use $\mathcal{S}$ in combination with $\mathcal{M}$ and $\mathcal{D}$ to create four kinds of negative bags. Differently to positive sampling, negatives can be constructed across the full minibatch rather than solely from the current (subtitled video, dictionary) pairing. We first anchor negatives bags on foreground segments: 

\begin{align}
 \mathcal{N}_{\text{watch,read,lookup}}^{\text{bags(seg-fore)}} = \{\{i\} \times B_i: (x_i^\mathcal{M}, v_i^\mathcal{M}) \in \mathcal{M}, (x_j^\mathcal{D}, v_j^\mathcal{D}) \in \mathcal{D},  \\ \nonumber B_i = \{j : v_j^\mathcal{D} \neq v_i^\mathcal{M} \}\}
\end{align}

\noindent so that they contain pairs between a given foreground segment and all available dictionary videos whose label does not match the segment (visualised in Fig.~\ref{app:fig:sampling_mil2nce} (i), both rows). We next anchor on the foreground dictionary videos:

\begin{align}
    \mathcal{N}_{\text{watch,read,lookup}}^{\text{bags(dict-fore)}} = \{A_i \times B_i: (x_i^\mathcal{D}, v_i^\mathcal{D}) \in \mathcal{D}, A_i = \{j : v_i^\mathcal{D} \in \text{Tokenize}(s_k),\\ \nonumber (x_k, s_k) \in \mathcal{S}, x_j \subseteq x_k, (x_j, v_j) \notin \mathcal{M}\} \cup \{(x_m, v_m) \in \mathcal{M}, v_m \neq v_i \}, \\ \nonumber B_i = \{l: v_l^\mathcal{D} = v_i^\mathcal{D}\}\}
\end{align}

\noindent comprising of pairings between the dictionary foreground set and segments within the minibatch that are either labelled with a different word, or can be excluded as a potential match through the subtitles (Fig.~\ref{app:fig:sampling_mil2nce} (ii), both rows).  Next, we anchor on the background continuous segments:
\nopagebreak
\begin{align}
 \mathcal{N}_{\text{watch,read,lookup}}^{\text{bags(seg-back)}} = \{\{i\} \times B_i: \exists (x_k, s_k) \in \mathcal{S}, x_i \subseteq x_k, (x_j^\mathcal{D}, v_j^\mathcal{D}) \in \mathcal{D}, \\ \nonumber B_i = \{j : v_j^\mathcal{D} \notin \text{Tokenize}(s_k) \}\}
\end{align}

\noindent which amounts to the pairings between each background segment and the set of dictionary videos which do not correspond to any of the words in the background subtitles (Fig.~\ref{app:fig:sampling_mil2nce} (iii), both rows). The fourth negative bag set construction anchors on the background dictionaries:

\begin{align}
    \mathcal{N}_{\text{watch,read,lookup}}^{\text{bags(dict-back)}} = \{A_i \times B_i: (x_i^\mathcal{D}, v_i^\mathcal{D}) \in \mathcal{D}, A_i = \{j : v_i^\mathcal{D} \notin \text{Tokenize}(s_k),\\ \nonumber (x_k, s_k) \in \mathcal{S}, x_j \subseteq x_k, (x_j, v_j) \notin \mathcal{M}\} \cup \{(x_m, v_m) \in \mathcal{M}, v_m \neq v_i \}, \\ \nonumber B_i = \{l: v_l^\mathcal{D} = v_i^\mathcal{D}\}\}
\end{align}

\noindent and thus the pairings arise between dictionary examples for a background segment and its corresponding foreground segment, as well all segments from other batch elements (Fig.~\ref{app:fig:sampling_mil2nce} (iv), both rows).  These four sets of bags are combined to form the full negative bag set:

\begin{align}
\mathcal{N}_{\text{watch,read,lookup}}^{\text{bags}} =
\mathcal{N}_{\text{watch,read,lookup}}^{\text{bags(seg-fore)}}
\cup \mathcal{N}_{\text{watch,read,lookup}}^{\text{bags(seg-dict)}} \\ \nonumber
\cup \, \mathcal{N}_{\text{watch,read,lookup}}^{\text{bags(seg-back)}}
\cup \mathcal{N}_{\text{watch,read,lookup}}^{\text{bags(dict-back)}}.
\end{align}

\noindent In the main paper, these bag formulations are used through Eqn. (1) (the MIL-NCE loss function) to guide learning.  Concretely, the \textit{Watch-Lookup} framework defines positive and negative bags via $\mathcal{P}^{\text{bags}} =\mathcal{P}_{\text{watch,lookup}}^{\text{bags}}$, $\mathcal{N}^{\text{bags}} =\mathcal{N}_{\text{watch,lookup}}^{\text{bags}}$ and the \textit{Watch-Read-Lookup} formulation instead defines the positive and negative bags via  $\mathcal{P}^{\text{bags}} =\mathcal{P}_{\text{watch,read,lookup}}^{\text{bags}}$, $\mathcal{N}^{\text{bags}} =\mathcal{N}_{\text{watch,read,lookup}}^{\text{bags}}$.

\subsection{Infrastructure}\label{app:subsec:infra}

The I3D trunk \bslonek{} pretraining experiments were performed with four Nvidia M40 graphics cards and took 2-3 days to complete. After freezing the I3D trunk, training the parameters of the MLP with the \textit{Watch-Read-Lookup} framework took approximately two hours on a single Nvidia M40 graphics card.

\end{document}